\documentclass{article}

\usepackage{arxiv}
\usepackage[square,numbers]{natbib}
\usepackage[utf8]{inputenc} 
\usepackage[T1]{fontenc}    
\usepackage{hyperref}       
\usepackage{url}            
\usepackage{booktabs}       
\usepackage{amsfonts}       
\usepackage{nicefrac}       
\usepackage{microtype}      
\usepackage{lipsum}		
\usepackage{graphicx}
\usepackage{doi}
\usepackage{tabularx}
\usepackage{graphicx}
\usepackage{subcaption}
\usepackage{amsmath}
\usepackage{mdframed}
\usepackage{longtable}
\newcolumntype{P}[1]{>{\raggedright\arraybackslash}p{#1}}
\usepackage{comment}
\usepackage{csquotes}
\usepackage[table]{xcolor}
\usepackage{multirow}
\usepackage{array}
\usepackage{authblk}
\usepackage{orcidlink}

\usepackage{listings}
\usepackage{soul}
\sethlcolor{white}
\soulregister\texttt7
\soulregister\ref7
\soulregister\cite7

\usepackage{xcolor}

\usepackage{fancyvrb}

\usepackage{textcomp}
\newcolumntype{Y}{>{\raggedright\arraybackslash}X}

\title{Resource Constraints and Performance in Agentic AI Systems}
\author[1]{Amaz  Salman}
\author[2]{Malka Halgamuge \orcidlink{0000-0001-9994-3778}}
\author[1]{Teo Susnjak\orcidlink{0000-0001-9416-1435}}

\affil[1]{School of Mathematical and Computational Sciences, Massey University, Auckland, New Zealand}
\affil[2]{Department of Information Systems and Business Analytics, RMIT University, Melbourne, VIC, Australia}

\renewcommand{\shorttitle}{Resource Constraints and Performance in Agentic AI Systems}

\begin{document}
\maketitle

\begin{abstract}
Progress toward more autonomous AI increasingly depends on agentic systems that combine a language model with tools, memory, state management, and multi-step execution. These mechanisms shape both task capability and operational burden. We compare OpenClaw and NanoBot as complete agentic systems using a paired primary benchmark and a more detailed instrumented subset of paired prompts. In the primary benchmark, the rate of full task completion was 31\% for OpenClaw and 25\% for NanoBot, a six-percentage-point difference with a 95\% task-bootstrap interval from $-3$ to 15 percentage points, providing no statistically established full-completion advantage for either system. In the instrumented layer, both systems achieved 26\% full completion, while NanoBot reached at least partial completion on 43\% of prompts compared with 26\% for OpenClaw. OpenClaw took longer on 83\% of prompts and had a higher recorded peak-memory value on every prompt, with geometric mean ratios of 2.98 for wall time and 19.44 for peak memory. Among the ten detailed-layer prompts on which at least one system achieved partial or full completion, NanoBot weakly dominated on eight; across all 23 prompts, however, ten of its eighteen dominance cases were cheaper joint failures. Outcome labels differ across the two evidence layers, showing why agent-system evaluation should connect capability and resource measurements to attempt-level execution and scoring provenance. These findings show that progress toward more autonomous AI should be evaluated through verified task completion, observed resource use and records linking each result to the execution that produced it.
\end{abstract}


\section{Introduction}

Large language models (LLMs) have moved agent research away from single-response generation and toward systems that plan, use tools, inspect intermediate results, and continue acting until a task reaches a verifiable end state. \hl{As a result, their evaluation must shift from the language model alone to the complete agentic system,} because the software harness shapes the action space, memory path, tool interface, orchestration policy, recovery logic, and operational footprint.

This system-level perspective is particularly important because autonomous action exposes weaknesses that one-shot language evaluation can hide. A language model may propose an action without maintaining a grounded and calibrated representation of the environment state, the action's preconditions, or its downstream effects. Training on linguistic form can encode broad regularities and world knowledge, but it does not by itself establish the connection between symbols, observations, actions, and consequences required for grounded understanding \cite{bender2020climbing}. Work on autonomous machine intelligence therefore treats predictive world models as a core component for reasoning across states and planning over several time scales \cite{lecun2022path}.

These limitations become operationally consequential when an agent must act across multiple steps. An agent may violate a constraint, consume an irreversible resource, lose access to a later action, or mistake an intermediate state for task completion. Errors can compound when later decisions depend on an earlier false state estimate. Benchmarks of planning and reasoning about change show that strong language models still struggle to produce executable plans, especially when success requires tracking preconditions and effects across unfamiliar instances \cite{valmeekam2023planning,valmeekam2023planbench}. Agent harnesses can add state tracking, search, retries, tool feedback, and verification. \hl{The empirical question is therefore how effectively an agentic system converts these mechanisms into task completion relative to its operational burden.}

This capability--cost question is also relevant to research on artificial general intelligence (AGI), although the present study does not claim to measure AGI. Progress toward broader, advanced machine intelligence depends on enabling paradigms, component capabilities, system architectures and evaluation methods. Agentic systems are relevant to that trajectory because they join planning, memory, tool use, environmental interaction and multi-step action. Each capacity addresses one part of the gap between language generation and autonomous problem solving. \citet{morris2024levels}
\hl{ distinguish performance, generality, and autonomy as separate dimensions of progress toward AGI, placing the evaluation of autonomous behaviour alongside capability breadth and depth}.

Within this broader agenda, harness evaluation addresses the enabling systems that support autonomous action. Future general-purpose autonomous systems will need to pursue goals across many state transitions, detect when the environment departs from their assumptions, and recover without uncontrolled resource growth. A harness can supply external state, search, verification, permissions, and tool feedback when the underlying model cannot maintain them with enough fidelity. \hl{We therefore evaluate this systems trade-off by pairing task outcomes with wall time and peak memory in a whole-system comparison.} Resource-bounded task completion provides one empirical step toward evaluating architectures that could support more general autonomous intelligence.

The same capability--cost question also matters for organisational deployment. Enterprises and public agencies obtain value from an agent only when it completes useful work with acceptable demands on infrastructure, human supervision, correction, and risk control. \hl{A benchmark pass establishes technical capability under specified conditions, while labour productivity and economic value require organisational measures.} Capability-cost evaluation supplies an intermediate layer: it measures whether \hl{an agentic system} reaches a defined outcome and records the operational burden required to do so.

\hl{Against this background, we evaluate OpenClaw and NanoBot as complete agentic systems. OpenClaw uses a service-oriented architecture with a broader orchestration surface, whereas NanoBot uses a smaller system with a flatter execution path. We compare their paired task outcomes and operational burden across two evidence layers, treating the complete system as the empirical unit.}
\hl{The analysis addresses three questions. First, how do the systems differ in full task completion and secondary ordinal outcomes across the primary paired benchmark? Second, how do they differ in wall time, peak memory, resource-bounded completion, and prompt-level dominance within the more detailed instrumented subset? Third, what does outcome disagreement between the layers reveal about the provenance required for robust agent-system comparison?}
\hl{The study contributes (1) a paired capability--cost evaluation of two complete agentic systems, (2) a cross-layer analysis of outcome concordance and run provenance, and a (3) set of design requirements for controlled and reproducible agent-system evaluation.} Collectively these contributions address an AGI-relevant systems requirement, namely that autonomous capability must remain useful under explicit resource and reproducibility constraints.
\section{ Background}

\subsection{Agent harnesses and autonomous action}\label{subsec:harnesses}

Autonomous-agent systems translate model outputs into actions. They route tool calls, maintain memory, manage execution state, retry failed steps, and terminate or escalate when a task stalls. Recent surveys and benchmark papers argue that agent evaluation must therefore measure the whole system rather than only the language model \cite{yehudai2026survey,mohammadi2025evaluation,emde2026maseval}. \citet{meng2026agent} use the term agent harness for the runtime system that combines an execution loop, tool registry, context manager, state store, lifecycle hooks, and evaluation interface. We adopt that term as the central object of study.

Tool integration expands an agent's functional reach, but it also adds orchestration work \cite{Zhang2026Generalizability,du2026Survey}. Tool calls require function and argument selection; their outputs, error signals, and environmental observations must be returned to the model for subsequent decisions; and actions through external systems create permission and security boundaries \cite{Zhang2026Generalizability,kim2026cost,he2025Emerged,yu2025survey}. Richer harnesses can improve task performance through iterative tool use, reflection, and alternative reasoning paths \cite{du2026Survey,kim2026cost}. These mechanisms can also increase model and tool invocations, latency, context and KV-cache memory demand, debugging and integration overhead, and failure and attack surfaces \cite{kim2026cost,du2026Survey,yu2025survey,he2025Emerged}.

\subsection{Position within the agent-system ecosystem}\label{subsec:agent-ecosystem}

The term agentic framework now covers systems with different users, action spaces, and execution boundaries. Agent-first development environments such as Codex, Claude Code, Google Antigravity, and OpenCode give agents access to repositories, editors, terminals, tests, browsers, and, in some cases parallel workspaces \cite{openai2026codex,anthropic2026claudecode,google2025antigravity,opencode2026docs}. Software-engineering platforms such as OpenHands and research systems such as SWE-agent expose related capabilities through reusable runtimes or benchmark-oriented agent--computer interfaces \cite{wang2025openhands,openhands2026sdk,yang2024sweagent}. These systems offer strong examples of tool-mediated planning and verification, but their software-development scope differs from the persistent, cross-domain assistance targeted by OpenClaw and NanoBot.

OpenClaw and Hermes Agent belong to a family of persistent general-purpose harnesses that connect models to tools, memory, communication channels, scheduled work, and delegated subagents \cite{openclaw2026,nous2026hermes}. A second cluster pursues a smaller runtime or architectural surface. NanoBot, NanoClaw, PicoClaw, and ZeroClaw make different choices about dependencies, isolation, deployment hardware, language runtime, and retained features \cite{hkuds2026nanobot,nanoclaw2026,picoclaw2026,zeroclaw2026}. 

Orchestration frameworks expose a different unit of analysis. LangGraph provides primitives for durable, stateful multi-step workflows; Microsoft Agent Framework develops orchestration concepts associated with AutoGen and Semantic Kernel; CrewAI organises agents, tools, and multi-agent workflows \cite{langgraph2026docs,microsoft2026agentframework,crewai2026docs}. These libraries help developers construct a harness, while OpenClaw and NanoBot are evaluated here as \hl{complete, runnable agentic systems}.
Table~\ref{tab:ecosystem-comparison} grounds the taxonomy in representative systems while comparing the architectural dimensions relevant to this study. The examples are illustrative rather than exhaustive.The family labels are an analytical synthesis informed by peer-reviewed work on agent architectures, frameworks, and agent--computer interfaces, together with the official system documentation cited above \cite{Zhang2026Generalizability,du2026Survey,wang2025openhands,yang2024sweagent}; they are not fixed product classes, and individual systems may cross category boundaries as their capabilities change. 

\begin{table*}[htbp]
\caption{Family-level comparison of contemporary agent systems. Examples are illustrative and non-exhaustive; bold type identifies the systems evaluated in this study. The synthesis uses the sources cited in Section~\ref{subsec:agent-ecosystem}, reviewed through 15 August 2026. Product capabilities and classifications remain version-dependent.}\label{tab:ecosystem-comparison}
\scriptsize
\setlength{\tabcolsep}{3.5pt}
\begin{tabularx}{\textwidth}{@{}p{0.145\textwidth}p{0.16\textwidth}p{0.19\textwidth}p{0.20\textwidth}Y@{}}
\toprule
System family & Illustrative systems & Typical autonomy surface & State and orchestration & Deployment, control, and evidence \\
\midrule
Agent-first development environments & Codex; Claude Code; Google Antigravity; OpenCode & Repository inspection, file editing, terminal execution, testing, and review; some systems add browser or workspace control & Session state, iterative tool loops, planning, verification, and optional delegation or parallel work & Local or cloud-backed development surfaces with approval and sandbox controls; evidence comes mainly from provider documentation \\
Software-engineering platforms and research agents & OpenHands; SWE-agent & Code, terminals, tests, issue descriptions, and benchmark environments & Explicit agent--computer interfaces, execution trajectories, retry policies, and task termination & Often containerised for evaluation; peer-reviewed benchmarks complement project documentation \\
Persistent general-purpose harnesses & \textbf{OpenClaw}; Hermes Agent & Messaging, browsers, APIs, schedules, communication channels, and cross-domain tools & Persistent memory, background tasks, delegation, channel routing, and long-lived services & Self-hosted or hybrid operation with system-specific permission models; independent comparative evidence remains limited \\
Lightweight harnesses and reimplementations & \textbf{NanoBot}; NanoClaw; PicoClaw; ZeroClaw & Selected cross-domain tools under a reduced dependency or runtime surface & Compact execution loops and selective memory, scheduling, or delegation features & Commonly self-hosted and intended for constrained deployment; evidence comes mainly from repositories and official documentation \\
Orchestration SDKs and graph frameworks & LangGraph; Microsoft Agent Framework; CrewAI & Developer-defined tools, agents, workflows, and application interfaces & Explicit state graphs, roles, hand-offs, checkpoints, retries, and single-agent or multi-agent coordination & Embedded within applications rather than deployed as end-user agents; research often evaluates primitives rather than complete products \\
\bottomrule
\end{tabularx}
\end{table*}

The comparison locates OpenClaw and NanoBot within adjacent architectural families without treating the family labels as fixed product identities. A harness determines how an agent records environment state, carries constraints across tool calls, checks action preconditions, and verifies whether an intervention produced the intended change. These functions become critical when the foundation model cannot maintain a grounded representation of the environment or predict action consequences with sufficient reliability. Thus we compare how OpenClaw and NanoBot maintain state, carry constraints across tool calls, and verify action outcomes.

\subsection{Grounding, world models, and prospective action}\label{subsec:world-models}

Autonomous multi-step action requires a system to identify its current environment state, represent temporal and causal dependencies, test action preconditions, and estimate possible successor states. We use \emph{world model} for the internal or external representation that supports these functions. Grounding connects that representation to observations, available actions, and their effects, allowing an agent to assess likely consequences before it acts. This requirement links world-model research directly to harness design, namely, external memory, state tracking, search, and post-action verification, which can compensate for weaknesses in the model that generates the action proposals.

Language modelling complicates this requirement. Text prediction can encode factual associations and plausible continuations, but it does not by itself establish a dependable connection among symbols, observations, actions, and consequences. \citet{bender2020climbing} distinguish linguistic form from grounding in communicative intent and the external world. \citet{lecun2022path} therefore places a configurable predictive world model at the centre of an architecture for autonomous machine intelligence. For agentic systems, the practical question is whether the combined model and harness can preserve state and anticipate consequences well enough to guide action when tasks depart from familiar linguistic patterns.

Planning evidence exposes the practical gap. PlanBench and related controlled studies report substantial deficits in tracking state transitions, preconditions, goal reachability, and executable plans \cite{valmeekam2023planbench,valmeekam2023planning}. CLadder evaluates associational, interventional, and counterfactual queries against formal causal graphs and finds the task challenging for language models even when a specialised chain-of-thought prompt improves performance \cite{jin2023cladder}. Performance also declines as temporal and constraint-satisfaction demands increase \cite{wang2024tram,chen2025lr2bench}. Chain-of-thought and repeated inference can help, but do not guarantee sound search or out-of-distribution constraint satisfaction \cite{stechly2024thoughtlessness}.

Explicit search, symbolic planners, simulators, and external verifiers can partly compensate. Reasoning via Planning uses Monte Carlo tree search and an LLM-based transition model to explore alternatives \cite{hao2023reasoning}; verifier feedback can also improve subsequent proposals \cite{valmeekam2023planning}. These mechanisms motivate harnesses as cognitive scaffolding while sharpening the systems trade-off: mitigation adds calls, latency, memory demand, and failure paths.

\subsection{Evaluation beyond final-answer success}\label{subsec:evaluation}

Benchmarking autonomous agents differs from benchmarking one-shot text generation. The evaluator must judge whether a system selected useful intermediate steps, used tools correctly, recovered from errors, respected constraints, and reached an externally checkable end state. Work on WebArena, WebArena Verified, OSWorld, and SWE-bench Verified reflects this shift toward executable environments and stronger evaluation protocols \cite{zhou2024webarena,hattami2025webarena,xie2024osworld,openai2024swebench}.

\hl{General-purpose and multi-environment benchmarks broaden this evaluation target. GAIA combines reasoning, multimodal information, web access, and tool use in questions designed for general AI assistants, while AgentBench evaluates reasoning and decision making across eight interactive environments} \cite{mialon2024gaia,liu2024agentbench}. \citet{yao2024taubench}'s \ensuremath{\tau}\hl{-bench adds end-state verification in tool-mediated domains and uses repeated trials to measure behavioural consistency}. Meanwhile, MASEval argues that frameworks and system topology can affect performance even when models and tasks remain fixed \cite{emde2026maseval}. While \citet{orogat2026understanding} report that multi-agent framework choices can change latency, throughput, planning accuracy, coordination success, and scalability. Recent work also catalogues architectural decisions in agent harnesses and examines executable code as a harness representation \cite{hu2026architectural,ning2026codeharness}. The claim that harnesses affect outcomes is therefore established. \hl{To that end, this study combines paired capability transitions, operational ratios, resource-bounded completion, prompt-level dominance, and a run-lineage audit in one empirical comparison.}

\subsection{Cost, reliability, and failure modes}\label{subsec:cost-failure}

Cost carries importance because autonomous agents may perform long sequences of tool-mediated steps. \citet{kim2026cost} show that dynamic reasoning and test-time scaling can increase resource use, latency variance, and infrastructure burden. \citet{rabanser2026towards} call for a science of AI agent reliability that measures repeated attempts and task-level variation rather than relying on isolated success rates. We therefore reserve the term reliability for repeated attempts under pinned conditions, not for a single benchmark pass. \citet{kapoor2025agents} \hl{ argue that agent evaluation should measure cost alongside accuracy and standardise benchmark practice to support reproducible comparisons}. \hl{We apply that capability--cost agenda to paired whole-system measurements and add resource-budget completion curves, prompt-level dominance, and an audit of evidence-layer lineage.}

Failure analysis also needs system-level categories. \citet{cemri2025multi} show that multi-agent systems fail through structural and coordination mechanisms as well as poor model answers. \citet{pandey2026evaluating}'s production-evaluation work and \citet{sun2026survey} 's security survey discuss production failure, drift, security, and guardrail risks in deployed agent settings. We use this work to distinguish infrastructure fragility from decision fragility in the current study.

Human-calibrated task horizons provide one connection between benchmark performance and the duration of expert work. \citet{kwa2025longtasks} estimate the task duration at which an agent reaches a stated success probability from repeated attempts on software, machine-learning, and cybersecurity tasks \cite{task-completion-time-horizons-of-frontier-ai-models}. \hl{Their method uses human-duration baselines and repeated attempts, whereas the short, medium, and long strata in this benchmark provide categorical groupings of execution demand.} METR\footnote{https://metr.org/time-horizons/} also reports lower performance on less structured tasks and cautions that time-horizon estimates do not measure independent operating time or establish that organizations can delegate all shorter tasks \cite{task-completion-time-horizons-of-frontier-ai-models,kwa2026limitations}.

Evidence from workplace use reinforces this distinction. In a randomised trial involving 16 experienced open-source developers and 246 tasks in repositories familiar to them, access to early-2025 AI tools increased measured completion time by 19\%, despite participants expecting a gain \cite{becker2025productivity}. A later METR study produced raw estimates consistent with speedups from newer agents, but participant selection and unreliable time measurement prevented a firm estimate of their magnitude \cite{metr2026productivityupdate}. These studies show that benchmark capability, task-level time reduction, and organisational value require separate measurements. Task substitution complicates inference because workers may shift toward activities that AI makes cheaper, so large speed gains on observed tasks need not yield gains of the same magnitude in produced value \cite{cunningham2026uplift}.

\section{Systems and Empirical Scope}\label{sec:systems}

\hl{The empirical dataset compares two agentic systems. The detailed execution data identify \texttt{openclaw-container-v3} and \texttt{nanobot-container-v3}, both operating in agent mode.}The primary benchmark records identify the systems by product name, but do not retain complete framework-version, model-setting, or attempt-level run information. Additional implementation records provide partial system and model details, as summarised in Table~\ref{tab:implementation-summary}. \hl{Table~\ref{tab:systems} summarises the systems and the scope of the comparison.}

\begin{table}[htbp]
\caption{\protect\hl{Agentic systems and empirical scope.}}\label{tab:systems}
\begin{tabularx}{\textwidth}{@{}p{0.22\textwidth}YY@{}}
\toprule
Dimension & OpenClaw & NanoBot \\
\midrule
\hl{System identifier} & \texttt{openclaw-container-v3} & \texttt{nanobot-container-v3} \\
\hl{System architecture} & \hl{Service-oriented system with a broader orchestration and runtime surface} & \hl{Lightweight system with a smaller runtime and execution surface} \\
Empirical unit & \hl{Complete agentic system} & \hl{Complete agentic system} \\
Observed measures & Outcome, time, CPU, memory, trace and call heuristics, termination & Outcome, time, CPU, memory, trace and call heuristics, termination \\
\hl{Analytical scope} & \hl{Whole-system outcomes} & \hl{Whole-system outcomes} \\
\bottomrule
\end{tabularx}
\end{table}

\hl{The system implementation, model, prompt, tools, permissions, runtime settings, and resource budgets jointly determine each observation. The paired comparison therefore estimates whole-system differences rather than the effects of individual components.}

\section{Data and Methods}\label{sec:study-design}

\subsection{Evidence layers and analytical grain}\label{subsec:exploratory-benchmark}

\hl{We analyse two evidence layers arising from our experiments. The primary layer contains 100 prompt-level observations for each system, with each prompt evaluated once by OpenClaw and once by NanoBot. Its variables include task category, benchmark horizon stratum (short, medium, or long), ordinal outcome score, startup latency, CPU usage, memory usage, and trace complexity. The detailed execution layer contains a subset of 23 prompts evaluated by both systems and adds prompt content, container version, execution mode, wall-clock duration, average and peak CPU usage, peak memory, trace and call heuristics, retries, termination reason, failure type, and manually assessed outcome. Prompt identifiers link the layers at task level, but the available records do not establish attempt-level linkage.}
The size of the selected benchmark has precedents and is comparable to component suites in peer-reviewed agent evaluations: \citet{yao2024taubench}'s \ensuremath{\tau}-bench contains 115 retail tasks and 50 airline tasks, while \citet{kwa2025longtasks}'s \hl{time-horizon evaluation uses 97 HCAST tasks, 66 software atomic-action tasks, and seven RE-Bench tasks.}

\hl{The detailed layer was constructed as a purposive, horizon-balanced subset comprising 7 short, 8 medium, and 8 long prompts. These prompts were selected for matched execution under tighter measurement conditions and to expose differences in workflow coordination, recovery, wall-clock duration, and memory use. The subset therefore complements the broader capability comparison with detailed operational evidence across all three execution horizons.}

\begin{table*}[htbp]
\caption{Provenance of the two analytical evidence layers.}\label{tab:provenance}
\fontsize{8pt}{10pt}\selectfont

\begin{tabularx}{\textwidth}{@{}p{0.22\textwidth}YY@{}}
\toprule
Provenance item & 100-prompt benchmark & 23-prompt detailed layer \\
\midrule
Pairing unit & \hl{Prompt identifier, one row per system} & \hl{Prompt identifier, one row per system} \\
\hl{System identifier} & Product label only & \texttt{openclaw-container-v3} and \texttt{nanobot-container-v3}, both in agent mode \\
Outcome record & Ordinal score and category metadata & Manually assessed ordinal outcome plus termination and sparse failure fields \\
Operational record & \hl{Startup-latency, CPU, memory, and trace fields; ten system-specific rows contain unexplained simultaneous zeros} & Wall time, CPU, peak memory, calls, retries, trace, and termination fields \\
Model, model settings, and prompts & \hl{Implementation record identifies the model and selected settings; prompt bank, identifiers, and categories recorded} & \hl{Implementation record identifies the model and selected settings; prompt text recorded} \\
Environment, tools, permissions, and budgets & \hl{Runtime and tool environment documented at system level; row-level settings absent} & \hl{Runtime and tool environment documented at system level; row-level settings absent} \\
Link between scores and executions & \hl{Prompt-level scores are available, but the executions that produced them are not identified} & \hl{Detailed executions are identified, but they cannot be matched to the executions that produced the corresponding scores in the 100-prompt dataset} \\
Subset construction & Not applicable & \hl{Purposive horizon-balanced selection for matched execution and detailed resource measurement} \\
\bottomrule
\end{tabularx}
\end{table*}
\hl{Table~\ref{tab:implementation-summary} summarises the system identity, model settings, runtime, benchmark execution, resource measurement, and scoring protocol. Appendix~\ref{app:implementation-provenance} provides the extended implementation record and identifies fields that were not recorded. Both systems used OpenAI gpt-4o-mini, while their inference settings and tool environments differed; these differences form part of the whole-system comparison.}
\begin{table*}[htbp]
\caption{\protect\hl{Key implementation and evaluation details for the two agentic systems.}}\label{tab:implementation-summary}
\fontsize{8pt}{10pt}\selectfont

\begin{tabularx}{\textwidth}{@{}p{0.22\textwidth}YY@{}}
\toprule
\hl{Dimension} & \hl{OpenClaw} & \hl{NanoBot} \\
\midrule
\hl{Evaluated release} & \hl{OpenClaw 2026.2.26-beta.1} & \hl{NanoBot v0.1.5} \\
\hl{Model} & \hl{OpenAI gpt-4o-mini} & \hl{OpenAI gpt-4o-mini} \\
\hl{Inference controls} & \hl{Temperature 0; top-p 1; context and output-token limits not recorded} & \hl{Temperature 0.1; provider-default top-p; 65,536-token context; 8,192-token output limit} \\
\hl{Runtime} & \hl{Docker Desktop/WSL2 Ubuntu; Node.js 22.22.2; npm 10.9.7} & \hl{Docker Desktop/WSL2 Ubuntu; Python 3.12.3} \\
\hl{Benchmark execution} & \hl{Shared 100-prompt suite; 30 short, 35 medium, and 35 long prompts; one scored attempt per prompt} & \hl{Same benchmark and attempt policy} \\
\hl{Tool environment} & \hl{Framework-native browser and execution tools; prepared benchmark services through Docker networking} & \hl{DuckDuckGo web search and execution tool; prepared benchmark services through Docker networking} \\
\hl{Resource measurement} & \hl{Harness wall time and container/runtime-level observed peak memory} & \hl{Same measurement boundaries and scope} \\
\hl{Outcome assessment} & \hl{Shared manual fail--partial--pass rubric; one non-blinded scorer} & \hl{Same rubric and scorer} \\
\hl{Execution period} & \hl{April--June 2026} & \hl{April--June 2026} \\
\bottomrule
\end{tabularx}
\end{table*}
\hl{The 23 prompts in the detailed execution layer also appear in the 100-prompt benchmark, with matching prompt metadata for both systems. Outcome scores differ across the layers: OpenClaw scores match on 15 of 23 prompts and NanoBot scores match on 13. We use the 100-prompt benchmark for the primary outcome analysis and analyse the 23-prompt execution data as a separate matched layer. We do not pool observations across layers or interpret the detailed data as repeated trials of the primary benchmark.}
\hl{A descriptive selection audit compares the same 23 prompt identifiers with the remaining 77 records in the primary layer. Horizon composition is similar: the subset contains 7 short, 8 medium, and 8 long prompts, compared with 23, 27, and 27 in the remainder. Category coverage differs. The subset omits Accuracy, Continuity, and Startup, contains the only Traceability prompt, and represents Auditability, Reliability, and Tool Use at higher shares than the remainder. Primary-layer outcome composition is also more favourable in the subset, most visibly for NanoBot. Appendix~\ref{app:subset-audit} reports the counts. The selection audit uses horizon, category, and outcome, the fields shared across the two layers.} \hl{The benchmark defines short-horizon prompts as seconds-scale tasks, medium-horizon prompts as minutes-scale tasks, and long-horizon prompts as extended multi-stage executions. These strata express intended execution demand rather than measured duration thresholds. Because task category, environmental requirements, and task difficulty also vary across the strata, horizon-stratified results are interpreted as associations rather than as estimates of the causal effect of execution length. Table~\ref{tab:provenance} distinguishes documented fields from unspecified provenance.}
\hl{Table~\ref{tab:metrics} defines the capability outcomes and operational measures used in the analysis.}Because inference settings, tool interfaces, and some execution details were not identical, the comparison estimates differences between the two recorded configurations and does not isolate the causal effect of harness architecture alone.

\begin{table}[htbp]
\caption{Core outcome definitions used in the empirical analysis.}\label{tab:metrics}
\begin{tabularx}{\textwidth}{@{}p{0.24\textwidth}Y@{}}
\toprule
Metric & Definition and analytical role \\
\midrule
\hl{Full completion} & \hl{The agentic system reaches the specified end state; primary capability outcome.} \\
Partial completion & The agent makes verifiable task-relevant progress without reaching the full end state. \\
Failure & The agent fails to make verifiable task-relevant progress or cannot execute the task. \\
Mean ordinal score & Mean of task scores with failure = 0, partial completion = 0.5, and full completion = 1; secondary summary. \\
Operational burden & \hl{Runtime resources consumed by the agentic system, including wall time, memory, CPU time, tokens, tool calls, and cost where available.} \\
\bottomrule
\end{tabularx}
\end{table}
\subsection{Data cleaning and quality controls}\label{subsec:evidence-recovery}

\hl{Before analysis, we corrected two category-label errors (Retieval to Retrieval and Continunity to Continuity), replaced underscores with spaces in category names, and standardised horizon labels to Short, Medium, and Long. We paired the OpenClaw and NanoBot records by prompt identifier and analysed the recorded scores both as outcome categories and on their numeric scale. Each source file contained one record per prompt identifier.}

\hl{We restricted the operational analysis to the detailed execution layer because the resource fields in the full benchmark could not be interpreted consistently. Ten full-benchmark records---nine for OpenClaw and one for NanoBot---contained simultaneous zeros for startup latency, CPU usage, memory use, and trace complexity; all ten recorded task failure. We treated these resource values as missing while retaining the corresponding completion outcomes. Excluding the affected prompt pairs produced the same substantive capability comparison (Appendix~\ref{app:subset-audit}).}
\hl{For the detailed execution layer, we converted source-reported peak-memory values from KB to MiB by dividing by 1024, treating the recorded unit as KiB. This conversion does not affect the system ratios, provided that both systems used the same unit and collection procedure. The records define the measure as container/runtime peak memory but do not identify the collector or sampling interval. We excluded CPU usage because the instrumentation did not define its denominator and recorded values exceeded 100\%.}

\hl{The detailed results dataset contains 30 failed executions: 17 for OpenClaw and 13 for NanoBot. A failure cause was recorded for five OpenClaw failures, all classified as framework crashes, and two NanoBot failures, both classified as hallucinations. Because causes were available for only seven failed executions, we did not compare failure types between the systems.}

\subsection{Statistical analysis}\label{subsec:statistics}

\hl{We analysed outcomes as paired observations indexed by prompt. Full task completion was the primary capability outcome because it does not impose equal spacing across failure, partial completion, and full completion. For paired binary outcomes, we report risk differences with 95\% percentile intervals obtained from 20,000 prompt-level bootstrap resamples that preserve the system pairing, together with exact McNemar tests.}These intervals describe sensitivity to the composition of the observed prompt set; they do not quantify run-to-run reliability or uncertainty for a population of future tasks. \hl{Partial-or-better completion was analysed as a secondary binary outcome. We also report outcome transitions, system wins, losses and ties, and mean ordinal scores. Paired mean-score differences use the same bootstrap procedure, while an exact two-sided sign test evaluates non-tied ordinal comparisons.}
\hl{For the detailed execution layer, right-skewed resource distributions are summarised using medians, interquartile ranges, and geometric mean OpenClaw-to-NanoBot ratios. Ratio intervals use the paired bootstrap, and exact sign tests assess the directional consistency of prompt-level resource differences. Resource-bounded completion curves show the proportion of prompts achieving at least partial completion without exceeding each observed wall-time or memory budget. A system weakly dominates when it achieves an equal or higher outcome with equal or lower wall time and peak memory, with at least one strict improvement. The principal dominance analysis excludes prompts on which both systems fail.}
\hl{Bootstrap intervals quantify sensitivity to the composition of the evaluated prompts rather than variation across repeated executions. Category-level analyses are exploratory: nine of the 17 categories contain fewer than five prompts, so we report paired mean differences and bootstrap intervals without multiplicity-adjusted hypothesis testing.}

\section{Results}\label{sec:results}

\subsection{Overall benchmark outcomes}\label{subsec:overall-outcomes}

\hl{OpenClaw reaches full completion on 31 of 100 prompts and NanoBot on 25}, a paired risk difference of 0.06 with a 95\% task-bootstrap interval from $-0.03$ to 0.15. Fourteen discordant prompts favour OpenClaw and eight favour NanoBot; the exact McNemar test gives $p=0.286$. Partial-or-better completion is 52\% and 47\%, with $p=0.551$. \hl{The primary estimate favours OpenClaw by six percentage points, with an interval that includes zero.}
The secondary ordinal mean is 0.415 for OpenClaw and 0.360 for NanoBot. The paired OpenClaw-minus-NanoBot difference is 0.055, with a 95\% task-bootstrap interval from $-0.040$ to 0.150. OpenClaw wins 29 paired ordinal comparisons, NanoBot wins 23, and 48 tie. The two-sided exact sign test over 52 non-tied pairs gives $p=0.488$.

\begin{figure}[htbp]
\centering
\includegraphics[width=0.78\textwidth]{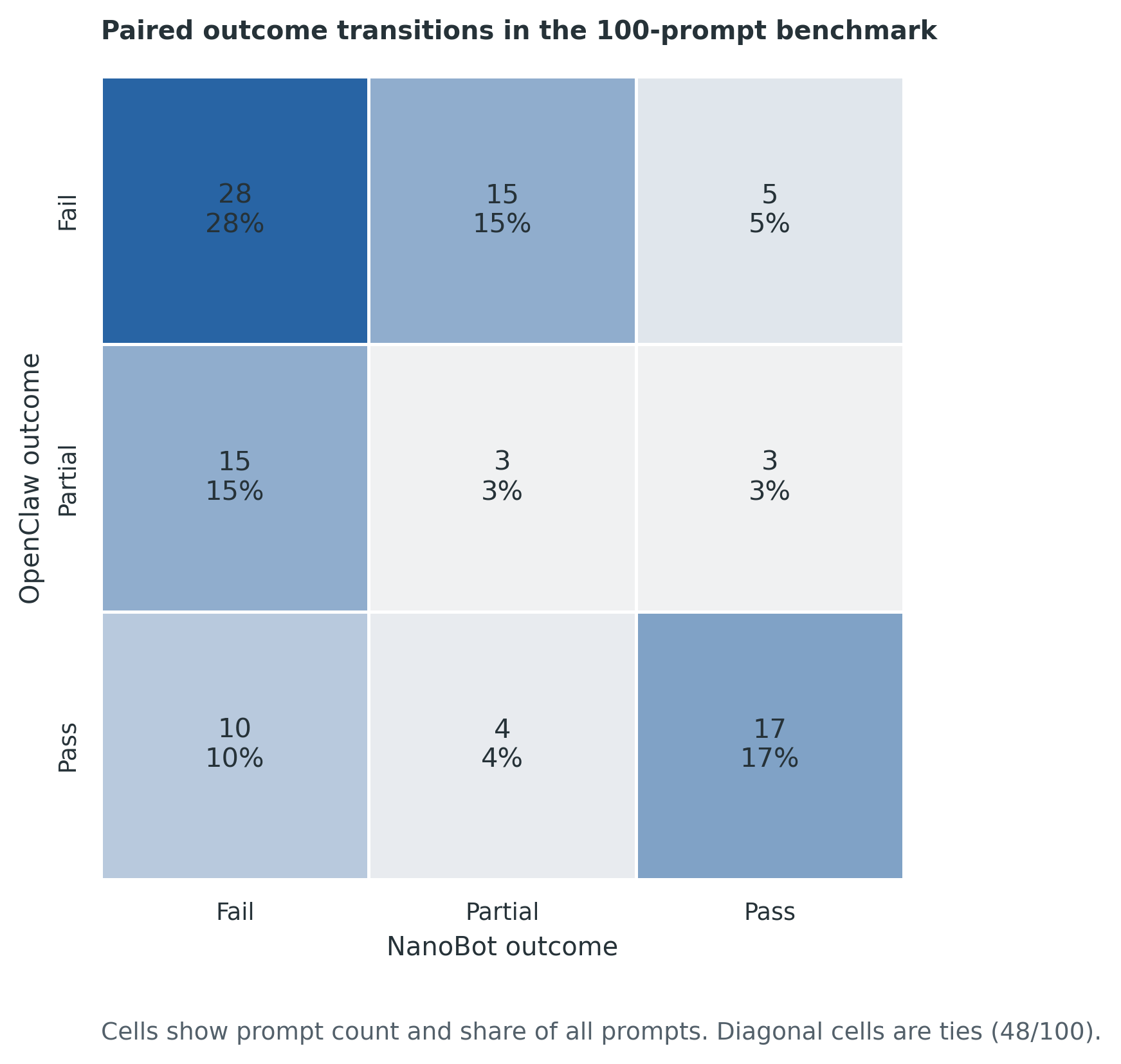}
\caption{Paired fail, partial, and pass transitions in the 100-prompt benchmark. Each cell gives a prompt count and its share of the full benchmark. Diagonal cells represent the 48 tied outcomes.}\label{fig:outcome-transitions}
\end{figure}

Figure~\ref{fig:outcome-transitions} shows the joint outcome structure hidden by separate aggregate bars. Both systems fail 28 prompts and reach full completion on 17. Fifteen OpenClaw failures become NanoBot partial completions, while 15 NanoBot failures become OpenClaw partial completions. OpenClaw converts ten NanoBot failures into full completions; NanoBot converts five OpenClaw failures into full completions. \hl{The remaining seven disagreements exchange full and partial outcomes.} The disagreement runs in both directions.
The transition structure confirms that the small aggregate differences mask substantial prompt-level disagreement. \hl{The comparison therefore reveals task-dependent differences between the systems rather than a uniform ordering.}

\begin{figure*}[htbp]
\centering
\includegraphics[width=1.0\textwidth]{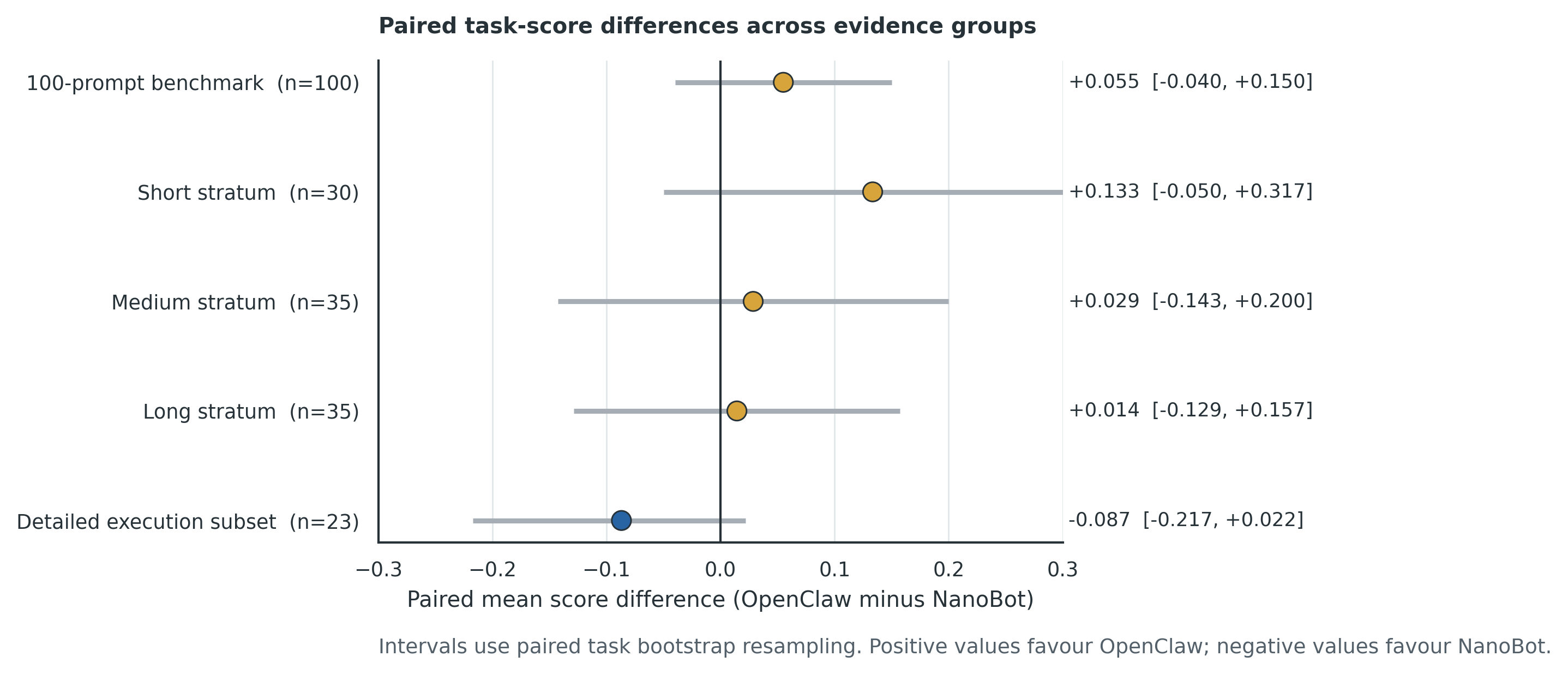}
\caption{Paired effects for the secondary ordinal task score across the primary benchmark, benchmark horizon strata, and detailed execution subset. Positive values favour OpenClaw; negative values favour NanoBot. Horizontal lines show 95\% paired task-bootstrap intervals. \protect\hl{The detailed execution interval represents its separate matched evidence layer.}}\label{fig:score-effects}
\end{figure*}

Figure~\ref{fig:score-effects} places the secondary ordinal effect beside the horizon strata and detailed execution subset. Each interval crosses zero. The estimates narrow from 0.133 in the short stratum to 0.014 in the long stratum, while the detailed execution estimate reverses direction to $-0.087$. \hl{The change in direction demonstrates that comparative performance depends on the evidence layer and prompt set.}

\subsection{Performance across benchmark horizon strata}\label{subsec:horizon-results}

\begin{figure}[htbp]
\centering
\includegraphics[width=1.0\textwidth]{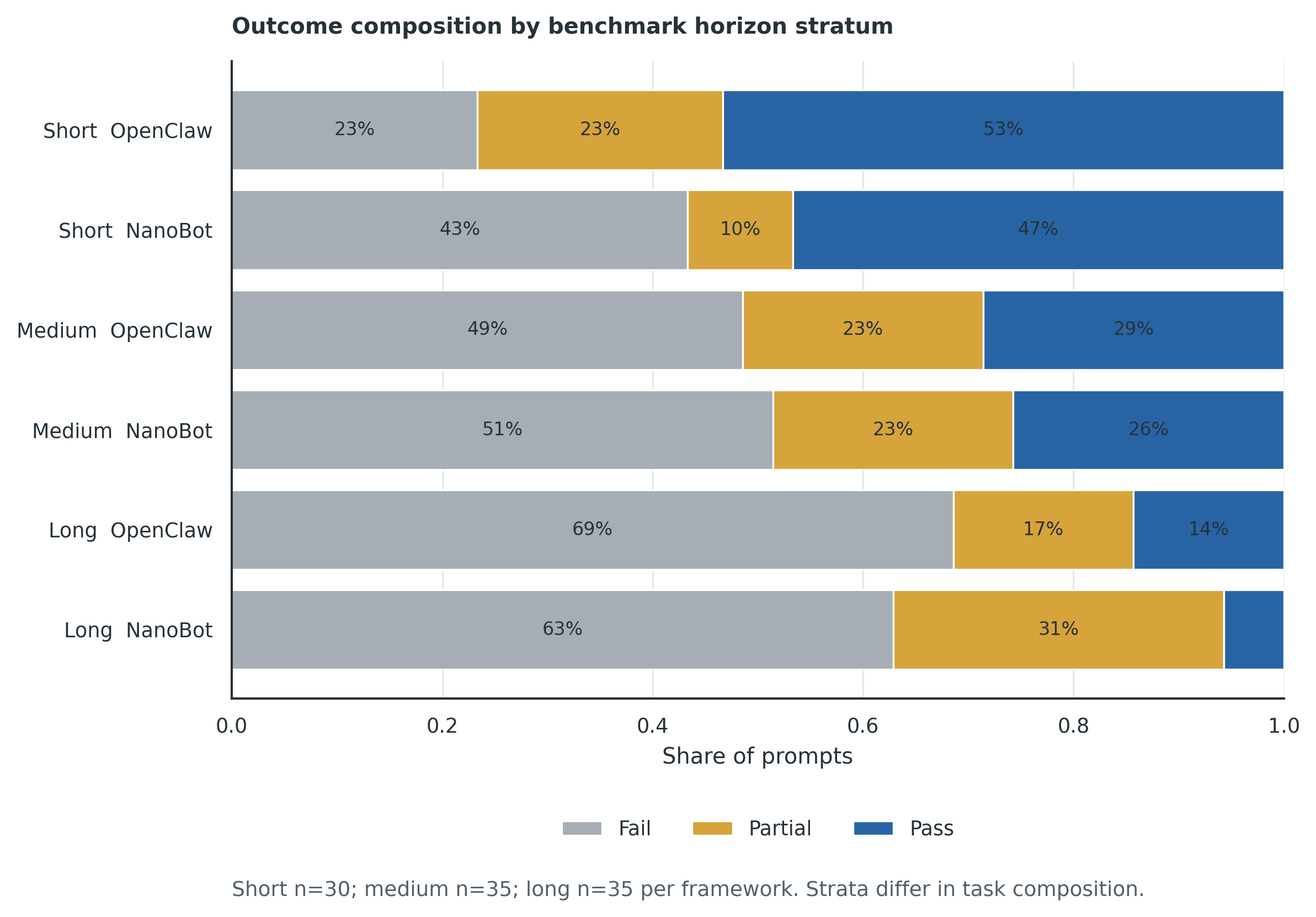}
\caption{Outcome composition by benchmark horizon stratum. \protect\hl{Each system contributes 30 short, 35 medium, and 35 long prompts.} The strata differ in task composition, so the figure reports association rather than a causal effect of execution horizon.}\label{fig:horizon-composition}
\end{figure}

Both systems record lower ordinal scores from the short to long benchmark strata. OpenClaw means fall from 0.650 to 0.400 and 0.229. NanoBot means fall from 0.517 to 0.371 and 0.214. Figure~\ref{fig:horizon-composition} shows that outcome composition provides a clearer account of the decline than the means alone. OpenClaw full completion falls from 53\% of short prompts to 14\% of long prompts, while its failure share rises from 23\% to 69\%. NanoBot full completion falls from 47\% to 6\%, while failure rises from 43\% to 63\%.
\hl{The paired system difference also contracts across the strata, but each interval remains wide. The stratified association combines execution depth with changes in category and environment demand.}

\subsection{Category and process patterns}\label{subsec:category-process}

\hl{The benchmark contains 17 categories, nine with fewer than five prompts. Among categories containing at least seven prompts, paired estimates favour OpenClaw in Retrieval, Stress, Workflow, Recovery, Auditability, and Continuity, and NanoBot in Memory; all corresponding intervals include zero. Larger contrasts appear in Tool Use, Accuracy, Tool Routing, Navigation, Orchestration, and Reliability, but each contains only three or four prompts. Because category also overlaps with horizon and environmental demand, these estimates identify task-dependent patterns rather than stable category-specific advantages. Figure~\ref{fig:category-effects} in the appendix reports the category estimates and prompt counts.}

\subsection{Operational burden}\label{subsec:operational-burden}

\begin{figure}[htbp]
\centering
\includegraphics[width=0.98\textwidth]{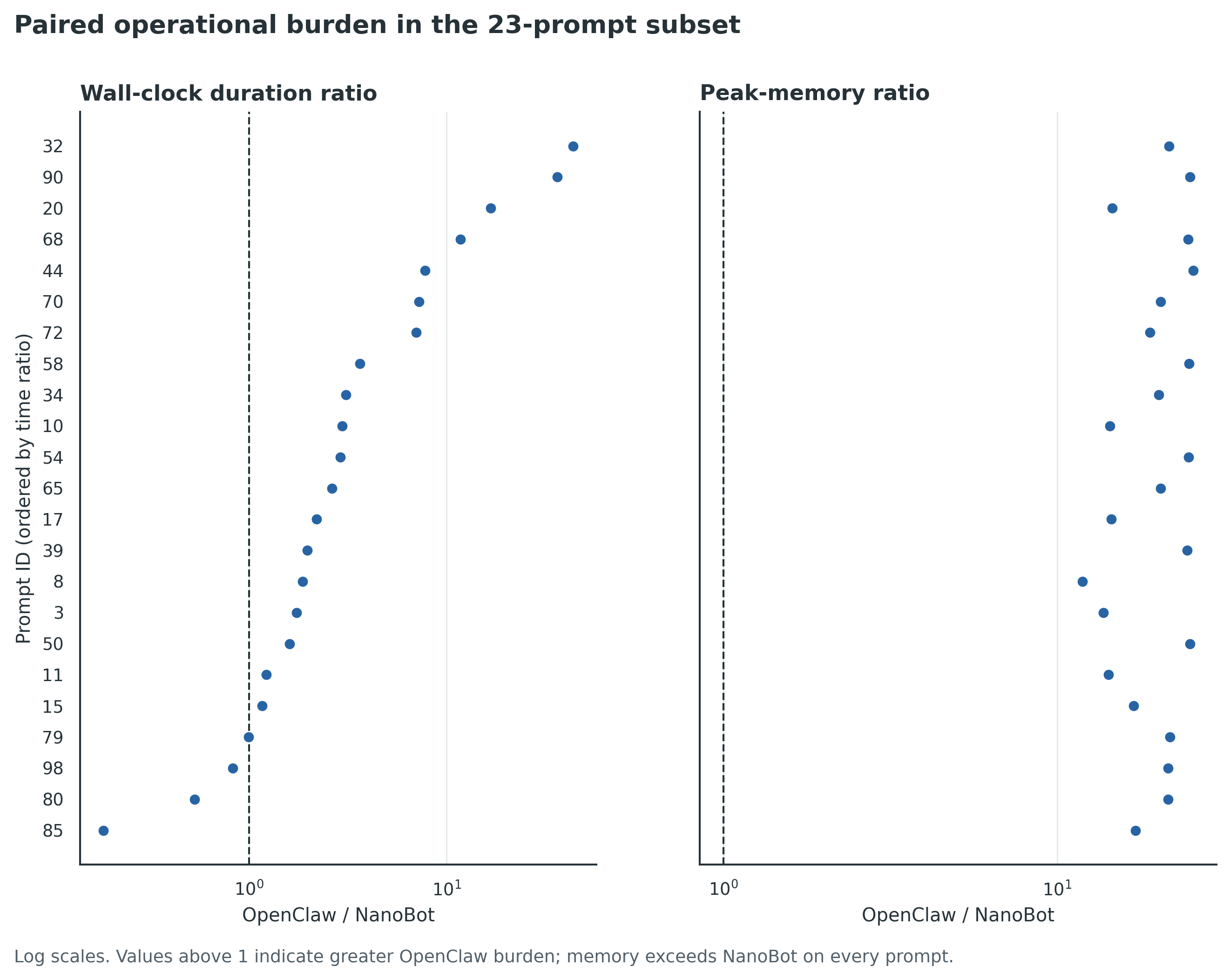}
\caption{Prompt-level OpenClaw-to-NanoBot ratios for wall-clock duration and peak memory in the detailed execution subset. Both panels use logarithmic axes, and the vertical reference line marks equal burden. Prompt identifiers follow the ordering of the wall-time ratio in both panels.}\label{fig:resource-ratios}
\end{figure}

\hl{The 23-prompt detailed execution layer contains matched wall-clock and peak-memory measurements.} OpenClaw has a median wall time of 34.063 seconds (interquartile range 13.040--70.754), compared with 10.446 seconds (6.528--17.212) for NanoBot. OpenClaw takes longer on 19 prompts and NanoBot on four. The geometric mean OpenClaw-to-NanoBot time ratio is 2.98, with a 95\% paired task-bootstrap interval from 1.78 to 5.06 and an exact directional $p=0.0026$. \hl{These estimates characterise the purposively selected, horizon-balanced detailed prompt set.}
Median peak memory is 2926.6 MiB for OpenClaw and 136.1 MiB for NanoBot under the source-unit assumption described in Section~\ref{subsec:evidence-recovery}. OpenClaw records higher peak memory for all 23 prompts. The geometric mean ratio is 19.44, with a \hl{95\% paired task-bootstrap interval} from 17.61 to 21.32 and an exact directional $p<0.001$. \hl{The paired ratios use the container/runtime-level measure recorded for both systems.} Figure~\ref{fig:resource-ratios} shows that these differences occur across prompt pairs rather than through one outlier.

Table~\ref{tab:paired-results} places these resource effects beside the outcome results. In the detailed execution subset, \hl{both systems achieve six full completions, and NanoBot achieves four additional partial completions.} The secondary ordinal mean therefore favours NanoBot by 0.087, with a \hl{95\% task-bootstrap interval} from $-0.217$ to 0.022. Eighteen prompts tie, four favour NanoBot, and one favours OpenClaw.

\begin{table*}[htbp]
\caption{Principal paired outcome and resource results. \protect\hl{Bootstrap intervals quantify prompt-resampling variation across one attempt per system--prompt pair.}}\label{tab:paired-results}
\fontsize{8pt}{10pt}\selectfont

\begin{tabularx}{\textwidth}{@{}p{0.25\textwidth}p{0.16\textwidth}p{0.16\textwidth}Y@{}}
\toprule
Measure & OpenClaw & NanoBot & Paired comparison \\
\midrule
100-prompt full completion (primary) & 31/100 & 25/100 & Risk difference 0.060; \hl{95\% task-bootstrap interval} $[-0.030,0.150]$; exact McNemar $p=0.286$ \\
100-prompt partial or better & 52/100 & 47/100 & Risk difference 0.050; exact McNemar $p=0.551$ \\
100-prompt mean ordinal score & 0.415 & 0.360 & Difference 0.055; \hl{95\% task-bootstrap interval} $[-0.040,0.150]$ \\
Detailed-layer full completion & 6/23 & 6/23 & Risk difference 0; exact McNemar $p=1.000$ \\
Detailed-layer partial or better & 6/23 & 10/23 & Risk difference $-0.174$; exact McNemar $p=0.125$ \\
Detailed-layer mean ordinal score & 0.261 & 0.348 & Difference $-0.087$; \hl{95\% task-bootstrap interval} $[-0.217,0.022]$ \\
Detailed-layer median wall time & 34.063 s & 10.446 s & Geometric ratio 2.98; \hl{95\% task-bootstrap interval} $[1.78,5.06]$ \\
Detailed-layer median peak memory & 2926.6 MiB & 136.1 MiB & Geometric ratio 19.44; \hl{95\% task-bootstrap interval} $[17.61,21.32]$ \\
\bottomrule
\end{tabularx}
\end{table*}

\subsection{Resource-bounded completion and dominance}\label{subsec:resource-bounded}

\begin{figure*}[htbp]
\centering
\includegraphics[width=0.96\textwidth]{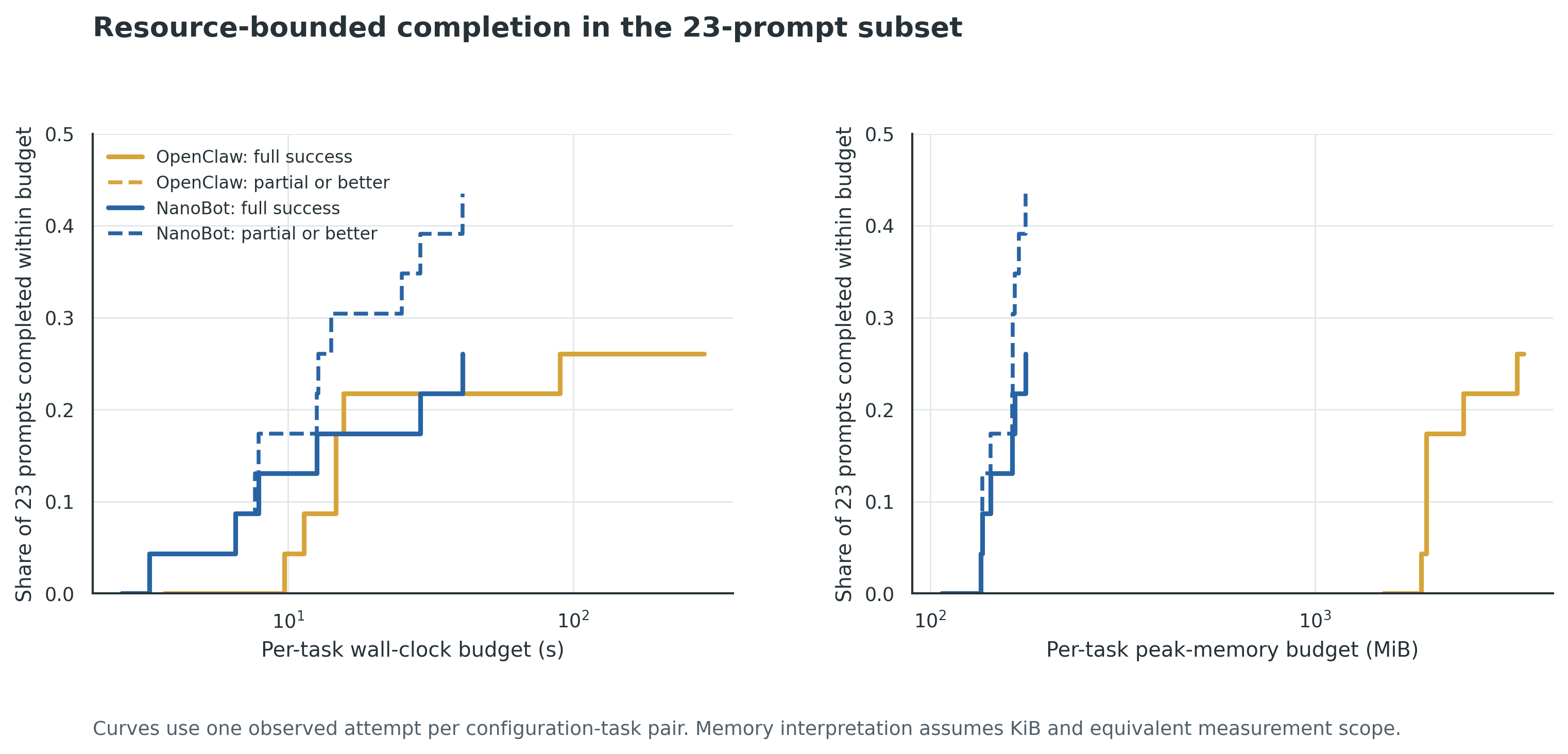}
\caption{Share of the 23 detailed execution prompts reaching full or partial-or-better completion within each observed per-task wall-time or peak-memory budget. \protect\hl{The curves use one attempt per system--prompt pair. The memory panel assumes that the recorded KB field represents KiB and that both systems used an equivalent measurement scope and procedure.}}\label{fig:budget-completion}
\end{figure*}

\hl{Figure~\ref{fig:budget-completion} shows how much of the 23-prompt set each system completes within an observed per-task budget.} NanoBot reaches three full completions within ten seconds, compared with one for OpenClaw. At 30 seconds, both reach five full task completions, while NanoBot also reaches partial-or-better completion on nine prompts compared with five for OpenClaw. Both reach six full completions at the largest observed time budgets. NanoBot reaches all six below 60 seconds; OpenClaw requires more than 60 seconds for its sixth.
\hl{The peak-memory curves separate the systems.} NanoBot reaches all six full completions and ten partial-or-better outcomes below 192 MiB. OpenClaw records no partial-or-better outcome below 1 GiB and reaches its sixth full completion at about 3.3 GiB. \hl{The curves quantify budget attainment within the evaluated prompt set.}

\begin{figure*}[htbp]
\centering
\includegraphics[width=1.0\textwidth]{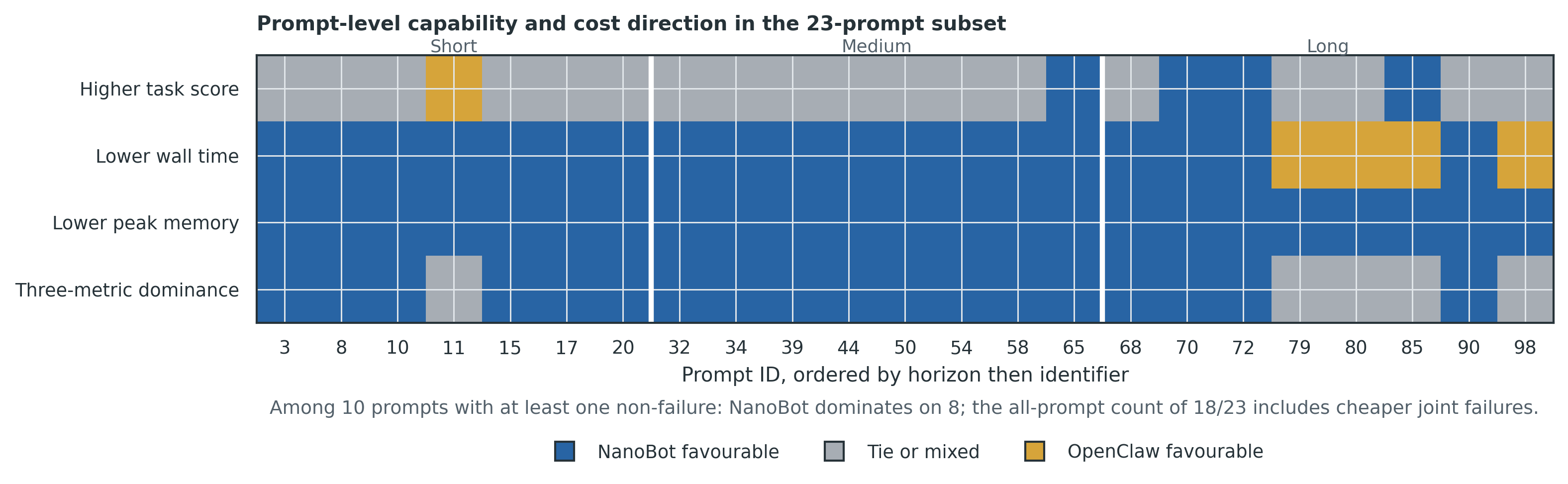}
\caption{Direction of task-score, wall-time, and peak-memory comparisons for each prompt in the detailed execution layer. Blue cells favour NanoBot, gold cells favour OpenClaw, and grey cells denote a score tie or mixed three-metric result. White separators mark the short, medium, and long benchmark strata.}\label{fig:dominance}
\end{figure*}

\hl{The principal dominance analysis excludes prompts on which both systems fail. Among the ten prompts with at least one partial or full completion, NanoBot weakly dominates on eight, OpenClaw on none, and two contain mixed trade-offs.}
\hl{Across all 23 prompts, NanoBot weakly dominates on 18. That count comprises three NanoBot score improvements, five joint passes obtained at lower observed cost, and ten joint failures terminated at lower observed cost. The ten cheaper joint failures contribute cost-only dominance; five prompts contain mixed trade-offs.} Figure~\ref{fig:dominance} displays the prompt-level directions, while Figure~\ref{fig:capability-cost-frontier} provides the aggregate capability-cost view in the appendix.

\begin{table}[htbp]
\caption{Sensitivity of three-metric weak dominance to joint failures.}\label{tab:dominance-sensitivity}
\begin{tabularx}{\textwidth}{@{}Yrrrr@{}}
\toprule
Prompt scope & $n$ & NanoBot & OpenClaw & Mixed \\
\midrule
All detailed-layer prompts & 23 & 18 & 0 & 5 \\
At least one non-failure & 10 & 8 & 0 & 2 \\
\bottomrule
\end{tabularx}
\end{table}

\subsection{Outcome consistency across datasets}\label{subsec:concordance}

The two datasets contain the same 23 prompts but do not always assign them the same outcome. OpenClaw's outcome differs on 8 prompts (34.8\%), and NanoBot's differs on 10 (43.5\%). Across these prompts, OpenClaw's mean score changes from 0.457 in the primary benchmark to 0.261 in the detailed dataset, while NanoBot's changes from 0.457 to 0.348. The records do not show whether these differences resulted from new executions, changes in the evaluation environment, or rescoring. We therefore analyse the datasets separately and interpret neither the matches nor the differences as evidence of repeated-run reliability. Figure~\ref{fig:cross-layer} shows the prompt-level changes in outcome.

\section{Discussion}\label{sec:discussion}

\hl{The primary benchmark does not establish a full-completion advantage for either system: OpenClaw completes six more prompts, but the interval includes zero. The detailed layer sharpens the comparison. Both systems record six full completions, while NanoBot records four more partial completions and uses less wall time and recorded peak memory. NanoBot also weakly dominates on eight of the ten prompts where at least one system makes verifiable progress.} Under the available measurement conditions, the OpenClaw configuration recorded substantially greater wall-time and peak-memory values without a demonstrated completion advantage. Because the configurations differed in inference settings, tools, and incompletely documented execution conditions, this result should not be interpreted as the causal cost of harness sophistication.

\hl{These results compare complete model--runtime--tool stacks. They show how the two systems behave on shared prompts and resource budgets, but they do not assign the difference to a particular harness component. This whole-system perspective is central to the study: tools, memory, state control, retries, and verification shape whether model outputs become successful actions and what those actions cost.}

\subsection{Implications for AGI evaluation}\label{subsec:agi-contribution}
The following implications concern evaluation design for systems relevant to future autonomous intelligence; they are not measurements of AGI, grounding, causal reasoning, or internal world-model quality. \hl{The findings connect agent-system evaluation to resource-bounded autonomy. A general-purpose autonomous system must sustain goals across multiple actions while controlling time, memory, cost, and failure risk. Harness mechanisms can support longer action sequences, but they also consume resources and create additional points of failure. Completion and operational burden should therefore be evaluated together.}

\hl{The detailed results show why this joint view matters. Equal numbers of full completions conceal differences of about threefold in wall time and nineteenfold in recorded peak memory. The full benchmark also shows failure becoming the dominant outcome for both systems in the long-horizon stratum. Because horizon overlaps with category and environmental demand, this pattern identifies sustained execution as a hypothesis for controlled testing rather than an isolated causal effect.}

\hl{The outcome changes across datasets add a reproducibility requirement. A prompt identifier alone cannot show whether a changed score arose from a new execution, a revised assessment, a software change, or a different environment state. Evaluations of autonomous capability therefore need repeated runs and records that connect each outcome to the execution that produced it.}

\hl{Harnesses may also compensate for weaknesses in model state tracking by recording observations, checking preconditions, exploring alternatives, and verifying action effects. The present data do not isolate these mechanisms or reveal either model's internal representation. Linked execution traces could support behavioural analysis of state errors, constraint violations, and missing verification while keeping claims focused on observable system behaviour.}

\subsection{Design requirements for reproducible agent-system evaluation}\label{sec:confirmatory}

\hl{Two findings translate directly into benchmark design. Both systems fail on 28 prompts, yet the records cannot separate task difficulty from unavailable prerequisites or shared model limitations. Scores also change across datasets without enough information to identify the cause. Table~\ref{tab:confirmatory} sets out the controls and records needed to resolve these ambiguities.}

\begin{table}[htbp]
\caption{Design requirements for reproducible agent-system evaluation.}\label{tab:confirmatory}
\begin{tabularx}{\textwidth}{@{}p{0.25\textwidth}Y@{}}
\toprule
Design element & Requirement \\
\midrule
Task structure & \hl{Matched task families with the same domain, objective, environment, and tools but different execution depths} \\
Repeated execution & \hl{Enough independent repeats of every system--task pair to achieve a pre-specified level of precision} \\
Experimental controls & \hl{Pinned model and system versions, tools, permissions, resource limits, containers, and reset state} \\
Run records & \hl{A unique execution identifier linking each prompt, output, trace, score, system configuration, and resource record} \\
Primary outcome & \hl{Verified binary completion against a pre-specified end state} \\
Operational measures & \hl{Wall time, memory, CPU time, tokens, API cost, tool calls, retries, and termination reason, with each collector and measurement boundary specified} \\
Assessment & \hl{System-blinded scoring where feasible, independent assessment, a fixed failure taxonomy, and reported agreement} \\
Analysis & \hl{Preserve system--prompt pairing, account for task families and repeated runs, and report effect intervals and run-to-run variation} \\
\bottomrule
\end{tabularx}
\end{table}

\hl{Matched task families would separate execution depth from task category and environment. Repeated runs would estimate reliability and resource variation, while linked records would show why scores or system rankings changed.}

\subsection{Implications for organisational deployment}\label{subsec:deployment}

\hl{The benchmark addresses technical capability and system efficiency, the first two levels in Table~\ref{tab:deployment-levels}. Organisational evaluation adds the human effort, implementation cost, service outcomes, and consequences that determine whether an agent creates value in practice.}

\begin{table*}[htbp]
\caption{Evaluation levels connecting agent benchmarks to organisational deployment. \protect\hl{The final column states the evidence included in this study.}}\label{tab:deployment-levels}
\fontsize{8pt}{10pt}\selectfont

\begin{tabularx}{\textwidth}{@{}p{0.14\textwidth}p{0.24\textwidth}p{0.32\textwidth}Y@{}}
\toprule
Level & Evaluation question & Representative measures & Coverage here \\
\midrule
Technical capability & \hl{Does the agentic system reach a pre-specified, externally checkable end state?} & Full and partial completion, constraint compliance, recovery, and failure & \hl{One completion outcome per system--prompt pair} \\
System efficiency & Which operational burden accompanies the outcome? & Wall time, memory, tokens, tool calls, monetary cost, retries, and failure exposure & \hl{Wall time and peak memory in the 23-prompt layer} \\
Organisational value & Does deployment improve work after all implementation and control demands are counted? & Human setup and supervision, correction and rework, integration and maintenance, decision quality, service outcomes, and expected error harm & \hl{A field-evaluation extension beyond the benchmark scope} \\
\bottomrule
\end{tabularx}
\end{table*}

\hl{Wall time and memory demand matter for capacity planning, but they do not establish labour savings or financial return. Deployment studies must also count infrastructure, supervision, correction, integration, maintenance, and the consequences of error.}

\section{Limitations}\label{sec:limitations}

\hl{The results apply to the evaluated OpenClaw and NanoBot model--runtime--tool stacks, both using gpt-4o-mini. The whole-system design does not isolate individual harness components, and results may differ with another model or tool environment. Each system--prompt pair was evaluated once, so the benchmark estimates comparative performance rather than repeated-run reliability. Horizon also overlaps with task category and environmental demand, while the purposive 23-prompt subset has a different category and outcome mix from the remaining prompts. Operational findings therefore apply to that detailed subset.}

\hl{Some implementation and data provenance remain incomplete, including exact repository commits, full configuration records, and the relationship between the two datasets. Several prompts depended on credentials, prepared files, authenticated sessions, or external services, and the records do not confirm equivalent access for every task. Outcomes may therefore include environmental as well as system differences.}
Completion labels were assigned by a single non-blinded assessor, and no independent inter-rater agreement was available. Borderline partial-completion decisions may therefore contain assessment subjectivity.
\hl{Operational comparisons are limited to wall time and recorded peak memory. The memory ratio assumes the same unit and collection procedure for both systems; CPU, token use, and monetary cost were not available for comparable analysis. The benchmark therefore measures technical capability and selected resource demands. It does not estimate organisational value or test internal model mechanisms.}

\section{Conclusion}\label{sec:conclusion}

As AI agents take on longer and more autonomous sequences of actions, evaluating them by task completion alone becomes increasingly inadequate. Their practical capability depends on \hl{the complete agentic system}, including the mechanisms that manage tools, memory, state, retries, and verification, together with the operational burden these mechanisms introduce.

\hl{We evaluated two open-source agentic systems built on the same foundation model as complete systems, pairing task outcomes with measured operational burden.} The benchmark did not provide statistically conclusive evidence of a difference in full-completion rates between the two recorded configurations. 
\hl{Operational burden separated the systems by roughly threefold in wall time and nineteenfold in recorded peak memory, with the lightweight system weakly dominating most prompts on which either system succeeded.} On this benchmark, the OpenClaw configuration recorded higher infrastructure use without a demonstrated completion advantage. The available data do not isolate whether this difference arose from harness architecture, inference settings, tool environments, or other configuration factors.
\hl{Finally, outcome labels for the same prompts changed across evidence layers for over a third of prompts, making run and scoring lineage a first-order requirement of agent evaluation rather than a bookkeeping detail. Evaluation of systems intended for general autonomous intelligence should therefore report capability jointly with resource budgets and attempt-level provenance; the proposed repeated-run protocol operationalizes these requirements.

}

\section*{Statements and Declarations}

\paragraph{Data and code availability}
\hl{After publication, the authors will make all datasets, results and code and the reported analysis results publicly available through a GitHub repository.}

\paragraph{Author contributions}
\hl{Amaz Salman: Methodology, Software, Investigation, Data curation, Formal analysis, Writing -- original draft, Visualization. Teo Susnjak: Conceptualization, Methodology, Formal analysis, Validation, Writing -- original draft, Writing -- review} \& \hl{editing, Visualization, Supervision. Malka N. Halgamuge: Validation, Writing -- original draft, Writing -- review} \& \hl{editing.}

\bibliographystyle{unsrtnat}
\bibliography{sn-bibliography}  

@inproceedings{bender2020climbing,
  title={Climbing towards {NLU}: On Meaning, Form, and Understanding in the Age of Data},
  author={Bender, Emily M. and Koller, Alexander},
  booktitle={Proceedings of the 58th Annual Meeting of the Association for Computational Linguistics},
  pages={5185--5198},
  year={2020},
  address={Online},
  publisher={Association for Computational Linguistics},
  url={https://aclanthology.org/2020.acl-main.463/},
  doi={10.18653/v1/2020.acl-main.463}
}

@article{lecun2022path,
  title={A Path Towards Autonomous Machine Intelligence},
  author={LeCun, Yann},
  journal={OpenReview preprint},
  year={2022},
  url={https://openreview.net/forum?id=BZ5a1r-kVsf}
}

@inproceedings{valmeekam2023planning,
  title={On the Planning Abilities of Large Language Models: A Critical Investigation},
  author={Valmeekam, Karthik and Marquez, Matthew and Sreedharan, Sarath and Kambhampati, Subbarao},
  booktitle={Advances in Neural Information Processing Systems},
  volume={36},
  pages={75993--76005},
  year={2023},
  doi={10.52202/075280-3320}
}

@inproceedings{valmeekam2023planbench,
  title={{PlanBench}: An Extensible Benchmark for Evaluating Large Language Models on Planning and Reasoning about Change},
  author={Valmeekam, Karthik and Marquez, Matthew and Olmo, Alberto and Sreedharan, Sarath and Kambhampati, Subbarao},
  booktitle={Advances in Neural Information Processing Systems},
  volume={36},
  pages={38975--38987},
  year={2023},
  doi={10.52202/075280-1693}
}

@inproceedings{morris2024levels,
  title={Position: Levels of {AGI} for Operationalizing Progress on the Path to {AGI}},
  author={Morris, Meredith Ringel and Sohl-Dickstein, Jascha and Fiedel, Noah and Warkentin, Tris and Dafoe, Allan and Faust, Aleksandra and Farabet, Clement and Legg, Shane},
  booktitle={Proceedings of the 41st International Conference on Machine Learning},
  series={Proceedings of Machine Learning Research},
  volume={235},
  pages={36308--36321},
  publisher={PMLR},
  year={2024},
  url={https://proceedings.mlr.press/v235/morris24b.html}
}

@inproceedings{yehudai2026survey,
  title={A Survey on Evaluation of {LLM}-based Agents},
  author={Yehudai, Asaf and Eden, Lilach and Li, Alan and Uziel, Guy and Zhao, Yilun and Bar-Haim, Roy and Cohan, Arman and Shmueli-Scheuer, Michal},
  booktitle={Findings of the Association for Computational Linguistics: ACL 2026},
  pages={26690--26714},
  year={2026},
  address={San Diego, California, United States},
  publisher={Association for Computational Linguistics},
  url={https://aclanthology.org/2026.findings-acl.1330/},
  doi={10.18653/v1/2026.findings-acl.1330}
}

@inproceedings{mohammadi2025evaluation,
  title={Evaluation and Benchmarking of {LLM} Agents: A Survey},
  author={Mohammadi, Mahmoud and Li, Yipeng and Lo, Jane and Yip, Wendy},
  booktitle={Proceedings of the 31st ACM SIGKDD Conference on Knowledge Discovery and Data Mining V.2},
  pages={6129--6139},
  year={2025},
  publisher={Association for Computing Machinery},
  doi={10.1145/3711896.3736570}
}

@inproceedings{emde2026maseval,
  title={{MASEval}: Extending Multi-Agent Evaluation from Models to Systems},
  author={Emde, Cornelius and Rubinstein, Alexander and Goel, Anmol and Heakl, Ahmed and Yun, Sangdoo and Oh, Seong Joon and Gubri, Martin},
  booktitle={Proceedings of the 64th Annual Meeting of the Association for Computational Linguistics (Volume 3: System Demonstrations)},
  pages={345--356},
  year={2026},
  address={San Diego, California, United States},
  publisher={Association for Computational Linguistics},
  url={https://aclanthology.org/2026.acl-demo.34/},
  doi={10.18653/v1/2026.acl-demo.34}
}

@misc{meng2026agent,
  title={Agent Harness for Large Language Model Agents: A Survey},
  author={Meng, Qianyu and Wang, Yanan and Chen, Liyi and Li, Yihang and Wu, Wei and Jiang, Wenyuan and Wang, Qimeng and Lu, Chengqiang and Gao, Yan and Wu, Yi and Hu, Yao},
  year={2026},
  howpublished={Preprints.org preprint},
  url={https://www.preprints.org/manuscript/202604.0428},
  note={Version 3, posted April 28, 2026}
}

@misc{openai2026codex,
  author={{OpenAI}},
  title={{Codex CLI}: Getting Started},
  year={2026},
  howpublished={OpenAI Help Center},
  url={https://help.openai.com/en/articles/11096431},
  note={Accessed: August 4, 2026}
}

@misc{anthropic2026claudecode,
  author={{Anthropic}},
  title={Claude Code Documentation},
  year={2026},
  url={https://docs.anthropic.com/en/docs/claude-code/getting-started},
  note={Accessed: August 4, 2026}
}

@misc{google2025antigravity,
  author={{Google Antigravity Team}},
  title={Build with Google Antigravity, Our New Agentic Development Platform},
  year={2025},
  month={November},
  url={https://developers.googleblog.com/en/build-with-google-antigravity-our-new-agentic-development-platform/},
  note={Accessed: August 4, 2026}
}

@misc{opencode2026docs,
  author={{OpenCode Team}},
  title={OpenCode: {AI} Coding Agent Built for the Terminal},
  year={2026},
  url={https://opencode.ai/docs},
  note={Accessed: August 4, 2026}
}

@misc{openhands2026sdk,
  author={{OpenHands}},
  title={OpenHands Software Agent SDK},
  year={2026},
  url={https://docs.openhands.dev/sdk/index},
  note={Accessed: August 4, 2026}
}

@inproceedings{wang2025openhands,
  title={{OpenHands}: An Open Platform for {AI} Software Developers as Generalist Agents},
  author={Wang, Xingyao and Li, Boxuan and Song, Yufan and Xu, Frank F. and Tang, Xiangru and Zhuge, Mingchen and Pan, Jiayi and Song, Yueqi and Li, Bowen and Singh, Jaskirat and Tran, Hoang H. and Li, Fuqiang and Ma, Ren and Zheng, Mingzhang and Qian, Bill and Shao, Yanjun and Muennighoff, Niklas and Zhang, Yizhe and Hui, Binyuan and Lin, Junyang and Brennan, Robert and Peng, Hao and Ji, Heng and Neubig, Graham},
  booktitle={International Conference on Learning Representations},
  year={2025},
  doi={10.48550/arXiv.2407.16741}
}

@inproceedings{yang2024sweagent,
  title={{SWE-agent}: Agent--Computer Interfaces Enable Automated Software Engineering},
  author={Yang, John and Jimenez, Carlos E. and Wettig, Alexander and Lieret, Kilian and Yao, Shunyu and Narasimhan, Karthik and Press, Ofir},
  booktitle={Advances in Neural Information Processing Systems},
  volume={37},
  year={2024},
  doi={10.52202/079017-1601}
}

@misc{openclaw2026,
  author={{OpenClaw Team}},
  title={OpenClaw: Open-source Autonomous Agentic {AI} Framework},
  year={2026},
  howpublished={GitHub repository},
  url={https://github.com/openclaw/openclaw},
  note={Accessed: March 25, 2026}
}

@misc{nous2026hermes,
  author={{Nous Research}},
  title={Hermes Agent},
  year={2026},
  url={https://hermes-agent.nousresearch.com/docs/},
  note={Accessed: August 4, 2026}
}

@misc{hkuds2026nanobot,
  author={{HKUDS Lab}},
  title={NanoBot: Ultra-lightweight Personal {AI} Assistant},
  year={2026},
  howpublished={GitHub repository},
  url={https://github.com/HKUDS/nanobot},
  note={Accessed: March 25, 2026}
}

@misc{nanoclaw2026,
  author={{NanoClaw Project}},
  title={NanoClaw: A Lightweight Container-Isolated Agent Harness},
  year={2026},
  howpublished={GitHub repository},
  url={https://github.com/nanocoai/nanoclaw},
  note={Accessed: August 4, 2026}
}

@misc{picoclaw2026,
  author={{Sipeed}},
  title={PicoClaw: Tiny, Fast, and Deployable Anywhere},
  year={2026},
  howpublished={GitHub repository},
  url={https://github.com/sipeed/picoclaw},
  note={Accessed: August 4, 2026}
}

@misc{zeroclaw2026,
  author={{ZeroClaw Labs}},
  title={ZeroClaw: Small Autonomous Personal-Assistant Infrastructure},
  year={2026},
  howpublished={GitHub repository},
  url={https://github.com/zeroclaw-labs/zeroclaw},
  note={Accessed: August 4, 2026}
}

@misc{langgraph2026docs,
  author={{LangChain}},
  title={LangGraph Overview},
  year={2026},
  url={https://langchain-ai.github.io/langgraph/},
  note={Accessed: August 13, 2026}
}

@misc{microsoft2026agentframework,
  author={{Microsoft}},
  title={Microsoft Agent Framework Overview},
  year={2026},
  url={https://learn.microsoft.com/en-us/agent-framework/overview/},
  note={Accessed: August 13, 2026}
}

@misc{crewai2026docs,
  author={{CrewAI}},
  title={CrewAI Documentation},
  year={2026},
  url={https://docs.crewai.com/},
  note={Accessed: August 13, 2026}
}

@inproceedings{jin2023cladder,
  author={Jin, Zhijing and Chen, Yuen and Leeb, Felix and Gresele, Luigi and Kamal, Ojasv and Lyu, Zhiheng and Blin, Kevin and Gonzalez Adauto, Fernando and Kleiman-Weiner, Max and Sachan, Mrinmaya and Sch{\"o}lkopf, Bernhard},
  title={{CL}adder: Assessing Causal Reasoning in Language Models},
  booktitle={Advances in Neural Information Processing Systems},
  volume={36},
  pages={31038--31065},
  year={2023},
  doi={10.52202/075280-1353}
}

@inproceedings{wang2024tram,
  title={{TRAM}: Benchmarking Temporal Reasoning for Large Language Models},
  author={Wang, Yuqing and Zhao, Yun},
  booktitle={Findings of the Association for Computational Linguistics: ACL 2024},
  pages={6389--6415},
  year={2024},
  address={Bangkok, Thailand},
  publisher={Association for Computational Linguistics},
  url={https://aclanthology.org/2024.findings-acl.382/},
  doi={10.18653/v1/2024.findings-acl.382}
}

@inproceedings{chen2025lr2bench,
  title={{LR$^2$Bench}: Evaluating Long-chain Reflective Reasoning Capabilities of Large Language Models via Constraint Satisfaction Problems},
  author={Chen, Jianghao and Wei, Zhenlin and Ren, Zhenjiang and Li, Ziyong and Zhang, Jiajun},
  booktitle={Findings of the Association for Computational Linguistics: ACL 2025},
  pages={6006--6032},
  year={2025},
  address={Vienna, Austria},
  publisher={Association for Computational Linguistics},
  url={https://aclanthology.org/2025.findings-acl.312/},
  doi={10.18653/v1/2025.findings-acl.312}
}

@inproceedings{stechly2024thoughtlessness,
  title={Chain of Thoughtlessness? An Analysis of Chain-of-Thought in Planning},
  author={Stechly, Kaya and Valmeekam, Karthik and Kambhampati, Subbarao},
  booktitle={Advances in Neural Information Processing Systems},
  volume={37},
  pages={29106--29141},
  year={2024},
  doi={10.52202/079017-0917}
}

@inproceedings{hao2023reasoning,
  title={Reasoning with Language Model is Planning with World Model},
  author={Hao, Shibo and Gu, Yi and Ma, Haodi and Hong, Joshua and Wang, Zhen and Wang, Daisy and Hu, Zhiting},
  booktitle={Proceedings of the 2023 Conference on Empirical Methods in Natural Language Processing},
  pages={8154--8173},
  year={2023},
  address={Singapore},
  publisher={Association for Computational Linguistics},
  url={https://aclanthology.org/2023.emnlp-main.507/},
  doi={10.18653/v1/2023.emnlp-main.507}
}

@inproceedings{zhou2024webarena,
  title={{WebArena}: A Realistic Web Environment for Building Autonomous Agents},
  author={Zhou, Shuyan and Xu, Frank F. and Zhu, Hao and Zhou, Xuhui and Lo, Robert and Sridhar, Abishek and Cheng, Xianyi and Ou, Tianyue and Bisk, Yonatan and Fried, Daniel and Alon, Uri and Neubig, Graham},
  booktitle={International Conference on Learning Representations},
  year={2024},
  url={https://openreview.net/forum?id=oKn9c6ytLx}
}

@inproceedings{hattami2025webarena,
  title={{WebArena} Verified: Reliable Evaluation for Web Agents},
  author={El Hattami, Amine and Thakkar, Megh and Chapados, Nicolas and Pal, Christopher},
  booktitle={Workshop on Scaling Environments for Agents},
  year={2025},
  url={https://openreview.net/forum?id=94tlGxmqkN}
}

@inproceedings{xie2024osworld,
  title={{OSWorld}: Benchmarking Multimodal Agents for Open-Ended Tasks in Real Computer Environments},
  author={Xie, Tianbao and Zhang, Danyang and Chen, Jixuan and Li, Xiaochuan and Zhao, Siheng and Cao, Ruisheng and Hua, Toh J. and Cheng, Zhoujun and Shin, Dongchan and Lei, Fangyu and Liu, Yitao and Xu, Yiheng and Zhou, Shuyan and Savarese, Silvio and Xiong, Caiming and Zhong, Victor and Yu, Tao},
  booktitle={Advances in Neural Information Processing Systems},
  volume={37},
  pages={52040--52094},
  year={2024},
  doi={10.52202/079017-1650}
}

@misc{openai2024swebench,
  author={{OpenAI}},
  title={Introducing {SWE}-bench Verified},
  year={2024},
  month={August},
  url={https://openai.com/index/introducing-swe-bench-verified/},
  note={Accessed: August 25, 2026}
}

@inproceedings{mialon2024gaia,
  author={Mialon, Gr{\'e}goire and Fourrier, Cl{\'e}mentine and Wolf, Thomas and LeCun, Yann and Scialom, Thomas},
  title={{GAIA}: A Benchmark for General {AI} Assistants},
  booktitle={The Twelfth International Conference on Learning Representations},
  year={2024},
  url={https://openreview.net/forum?id=fibxvahvs3}
}

@inproceedings{liu2024agentbench,
  author={Liu, Xiao and Yu, Hao and Zhang, Hanchen and Xu, Yifan and Lei, Xuanyu and Lai, Hanyu and Gu, Yu and Ding, Hangliang and Men, Kaiwen and Yang, Kejuan and Zhang, Shudan and Deng, Xiang and Zeng, Aohan and Du, Zhengxiao and Zhang, Chenhui and Shen, Sheng and Zhang, Tianjun and Su, Yu and Sun, Huan and Huang, Minlie and Dong, Yuxiao and Tang, Jie},
  title={{AgentBench}: Evaluating {LLM}s as Agents},
  booktitle={The Twelfth International Conference on Learning Representations},
  year={2024},
  url={https://openreview.net/forum?id=zAdUB0aCTQ}
}

@inproceedings{yao2024taubench,
  author={Yao, Shunyu and Shinn, Noah and Razavi, Pedram and Narasimhan, Karthik},
  title={{\ensuremath{\tau}}-bench: A Benchmark for Tool-Agent-User Interaction in Real-World Domains},
  booktitle={Proceedings of the Thirteenth International Conference on Learning Representations},
  year={2025},
  url={https://openreview.net/forum?id=roNSXZpUDN}
}

@article{orogat2026understanding,
  title={Understanding Multi-Agent {LLM} Frameworks: A Unified Benchmark and Experimental Analysis},
  author={Orogat, Abdelghny and Rostam, Ana and Mansour, Essam},
  journal={arXiv preprint arXiv:2602.03128},
  year={2026},
  doi={10.48550/arXiv.2602.03128}
}

@article{hu2026architectural,
  author={Hu, Wei},
  title={Architectural Design Decisions in {AI} Agent Harnesses},
  journal={arXiv preprint arXiv:2604.18071},
  year={2026},
  doi={10.48550/arXiv.2604.18071}
}

@article{ning2026codeharness,
  author={Ning, Xuying and Tieu, Katherine and Fu, Dongqi and others},
  title={Code as Agent Harness},
  journal={arXiv preprint arXiv:2605.18747},
  year={2026},
  doi={10.48550/arXiv.2605.18747}
}

@inproceedings{kim2026cost,
  author    = {Kim, Jiin and Shin, Byeongjun and Chung, Jinha and Rhu, Minsoo},
  booktitle = {2026 IEEE International Symposium on High Performance Computer Architecture (HPCA)},
  title     = {The Cost of Dynamic Reasoning: Demystifying AI Agents and Test-Time Scaling from an AI Infrastructure Perspective},
  year      = {2026},
  pages     = {1--16},
  publisher={IEEE Computer Society},

  doi       = {10.1109/HPCA68181.2026.11408569}
}

@inproceedings{rabanser2026towards,
  title={Towards a Science of {AI} Agent Reliability},
  author={Rabanser, Stephan and Kapoor, Sayash and Kirgis, Peter and Liu, Kangheng and Utpala, Saiteja and Narayanan, Arvind},
  booktitle={Proceedings of the 43rd International Conference on Machine Learning},
  year={2026},
  doi={10.48550/arXiv.2602.16666}
}

@article{kapoor2025agents,
  author={Kapoor, Sayash and Stroebl, Benedikt and Siegel, Zachary S. and Nadgir, Nitya and Narayanan, Arvind},
  title={{AI} Agents That Matter},
  journal={Transactions on Machine Learning Research},
  year={2025},
  url={https://openreview.net/forum?id=Zy4uFzMviZ}
}

@inproceedings{cemri2025multi,
  title={Why Do Multi-Agent {LLM} Systems Fail?},
  author={Cemri, Mert and Pan, Melissa Z. and Yang, Shuyi and Agrawal, Lakshya A. and Chopra, Bhavya and Tiwari, Rishabh and Keutzer, Kurt and Parameswaran, Aditya and Klein, Dan and Ramchandran, Kannan and Zaharia, Matei and Gonzalez, Joseph E. and Stoica, Ion},
  booktitle={Advances in Neural Information Processing Systems},
  volume={38},
  year={2025},
  doi={10.48550/arXiv.2503.13657}
}

@article{pandey2026evaluating,
  title={Evaluating Agentic {AI} in the Wild: Failure Modes, Drift Patterns, and a Production Evaluation Framework},
  author={Pandey, Mukund},
  journal={arXiv preprint arXiv:2605.01604},
  year={2026},
  doi={10.48550/arXiv.2605.01604}
}

@misc{sun2026survey,
  title={A Survey on the Unique Security of Autonomous and Collaborative {LLM} Agents: Threats, Defenses, and Futures},
  author={Sun, Yinggang and Yu, Haining and Jiang, Wei and Yu, Xiangzhan and Zhan, Dongyang and Wang, Lixu and Ren, Siyue and Sun, Yue and Zhu, Tianqing},
  year={2026},
  howpublished={Preprints.org preprint},
  url={https://www.preprints.org/manuscript/202602.1655},
  note={Version 3, posted August 6, 2026}
}

@inproceedings{kwa2025longtasks,
  author={Kwa, Thomas and West, Ben and Becker, Joel and Deng, Amy and Garcia, Katharyn and Hasin, Max and Jawhar, Sami and Kinniment, Megan and Rush, Nate and von Arx, Sydney and Bloom, Ryan and Broadley, Thomas and Du, Haoxing and Goodrich, Brian and Jurkovic, Nikola and Miles, Luke Harold and Nix, Seraphina and Lin, Tao and Painter, Chris and Parikh, Neev and Rein, David and Sato, Lucas Jun Koba and Wijk, Hjalmar and Ziegler, Daniel M. and Barnes, Elizabeth and Chan, Lawrence},
  title={Measuring {AI} Ability to Complete Long Software Tasks},
  booktitle={Advances in Neural Information Processing Systems},
  volume={38},
  year={2025},
  doi={10.52202/085713-3086},
  url={https://proceedings.neurips.cc/paper\_files/paper/2025/hash/85069585133c4c168c865e65d72e9775-Abstract-Conference.html}
}

@misc{task-completion-time-horizons-of-frontier-ai-models,
  author={{METR}},
  title={Task-Completion Time Horizons of Frontier {AI} Models},
  year={2026},
  month={March},
  url={https://metr.org/time-horizons/},
  note={Accessed: August 15, 2026}
}

@misc{kwa2026limitations,
  author={{METR}},
  title={Clarifying Limitations of Time Horizon},
  year={2026},
  month={January},
  howpublished={{METR} research note},
  url={https://metr.org/notes/2026-01-22-time-horizon-limitations/},
  note={Accessed: August 15, 2026}
}

@article{becker2025productivity,
  author={Becker, Joel and Rush, Nate and Barnes, Elizabeth and Rein, David},
  title={Measuring the Impact of Early-2025 {AI} on Experienced Open-Source Developer Productivity},
  journal={arXiv preprint arXiv:2507.09089},
  year={2025},
  doi={10.48550/arXiv.2507.09089}
}

@misc{cunningham2026uplift,
  author={Cunningham, Tom and Whitfill, Parker},
  title={Task Substitution and Uplift},
  year={2026},
  month={May},
  howpublished={{METR} research note},
  url={https://metr.org/blog/2026-05-08-task-substitution-and-uplift/},
  note={Accessed: August 15, 2026}
}

@misc{metr2026productivityupdate,
  author={{METR}},
  title={We Are Changing Our Developer Productivity Experiment Design},
  year={2026},
  month={February},
  howpublished={{METR} research update},
  url={https://metr.org/blog/2026-02-24-uplift-update/},
  note={Accessed: August 15, 2026}
}

@inproceedings{garg2025real,
  title={{REAL}: Benchmarking Autonomous Agents on Deterministic Simulations of Real Websites},
  author={Garg, Div and Caples, Diego and Draguns, Andis and Ravi, Nikil and Putta, Pranav and Garg, Naman and Hebbar, Prannay and Joo, Youngchul and Gu, Jindong and London, Charles and de Witt, Christian Schroeder and Motwani, Sumeet},
  booktitle={Advances in Neural Information Processing Systems},
  volume={38},
  year={2025},
  doi={10.52202/085713-4536}
}

@article{he2025Emerged,
author = {He, Feng and Zhu, Tianqing and Ye, Dayong and Liu, Bo and Zhou, Wanlei and Yu, Philip S.},
title = {The Emerged Security and Privacy of LLM Agent: A Survey with Case Studies},
year = {2025},
issue_date = {April 2026},
publisher = {Association for Computing Machinery},
address = {New York, NY, USA},
volume = {58},
number = {6},
issn = {0360-0300},
url = {https://doi.org/10.1145/3773080},
doi = {10.1145/3773080},
journal = {ACM Comput. Surv.},
month = dec,
articleno = {162},
numpages = {36}
}

@inproceedings{yu2025survey,
  title={A survey on trustworthy llm agents: Threats and countermeasures},
  author={Yu, Miao and Meng, Fanci and Zhou, Xinyun and Wang, Shilong and Mao, Junyuan and Pan, Linsey and Chen, Tianlong and Wang, Kun and Li, Xinfeng and Zhang, Yongfeng and others},
  booktitle={Proceedings of the 31st ACM SIGKDD Conference on Knowledge Discovery and Data Mining V. 2},
  pages={6216--6226},
  year={2025}
}

@article{du2026Survey,
author = {Du, Shangheng and Zhao, Jiabao and Shi, Jinxin and Xie, Zhentao and Jiang, Xin and Bai, Yanhong and He, Liang},
title = {A Survey on the Optimization of Large Language Model-based Agents},
year = {2026},
publisher = {Association for Computing Machinery},
address = {New York, NY, USA},
volume = {58},
number = {9},
issn = {0360-0300},
url = {https://doi.org/10.1145/3789261},
doi = {10.1145/3789261},
journal = {ACM Comput. Surv.},
month = feb,
articleno = {223},
numpages = {37}
}

@article{Zhang2026Generalizability,
author = {Zhang, Minxing and Yang, Yi and Xie, Roy and Dhingra, Bhuwan and Zhou, Shuyan and Pei, Jian},
title = {Generalizability of Large Language Model-Based Agents: A Comprehensive Survey},
year = {2026},
issue_date = {July 2026},
publisher = {Association for Computing Machinery},
address = {New York, NY, USA},
volume = {58},
number = {10},
issn = {0360-0300},
url = {https://doi.org/10.1145/3794858},
doi = {10.1145/3794858},
journal = {ACM Comput. Surv.},
month = apr,
articleno = {263},
numpages = {44}
}

\appendix

\clearpage

\section{Additional analyses}\label{app:additional-figures}

\hl{The appendix reports extended implementation provenance and supporting analyses of subset composition, category heterogeneity, capability--cost position, cross-layer outcome concordance, and simultaneous operational zeros.}

\subsection{Extended implementation and evaluation provenance}\label{app:implementation-provenance}

\hl{Table~\ref{tab:implementation-details} expands the implementation summary in Table~\ref{tab:implementation-summary}. It records the available system, model, benchmark, environment, measurement, scoring, and lineage details and marks information that was not recorded. The shared prompt suite drew on task patterns from REAL, WebArena and WebArena Verified, OSWorld, SWE-bench Verified, and MASEval} \cite{garg2025real,zhou2024webarena,hattami2025webarena,xie2024osworld,openai2024swebench,emde2026maseval}.

\begin{table*}[htbp]
\caption{\protect\hl{Extended implementation and evaluation provenance for the two agentic systems.}}\label{tab:implementation-details}
\fontsize{8pt}{10pt}\selectfont
\begin{tabularx}{\textwidth}{@{}p{0.12\textwidth}p{0.45\textwidth}Y@{}}
\toprule
\hl{Provenance item} & \hl{OpenClaw} & \hl{NanoBot} \\
\midrule
\hl{System release and repository commit} & \hl{OpenClaw 2026.2.26-beta.1; repository commit not recorded} & \hl{NanoBot v0.1.5; repository commit not recorded} \\
\hl{Container and runtime} & \hl{Docker container using the OpenClaw gateway/runtime through Docker Desktop and WSL2 Ubuntu; Node.js 22.22.2 and npm 10.9.7; image identifier and Ubuntu release not recorded} & \hl{Docker container \texttt{nanobot-benchmark} through Docker Desktop and WSL2 Ubuntu; Python 3.12.3; NanoBot executable \texttt{/tmp/nanobot-real/bin/nanobot}; image identifier and Ubuntu release not recorded} \\
\hl{Model provider and identifier} & \hl{OpenAI gpt-4o-mini; endpoint and dated model snapshot not recorded} & \hl{OpenAI gpt-4o-mini; endpoint and dated model snapshot not recorded} \\
\hl{Inference settings} & \hl{Temperature 0; top-p 1; context and maximum output-token limits not recorded} & \hl{Temperature 0.1; provider-default top-p; 65,536-token context; 8,192-token maximum output} \\
\hl{Benchmark and prompt protocol} & \hl{Shared 100-prompt suite from \texttt{AUTONOMOUS-AGENT-BENCHMARK/prompts.md}; 30 short, 35 medium, and 35 long prompts; prompts submitted individually; prompt-bank version, assignment procedure, system instructions, and delivery order not recorded} & \hl{Same benchmark, horizon allocation, and individual prompt-delivery method; prompt-bank version, assignment procedure, system instructions, and delivery order not recorded} \\
\hl{Horizon definitions} & \hl{Short: seconds-scale; medium: minutes-scale; long: extended multi-stage execution} & \hl{Same definitions} \\
\hl{Tools and prepared environment} & \hl{Framework-native browser and execution tools; OpenAI credentials; prepared benchmark services exposed through Docker networking; exact permissions and tool versions not recorded} & \hl{DuckDuckGo web search and execution tool; OpenAI credentials; prepared benchmark services and network access; workspace at \texttt{/.nanobot/workspace}; exact permissions and tool versions not recorded} \\
\hl{Host, scheduling, and resource controls} & \hl{Local workstation through Docker Desktop/WSL2; host CPU, RAM, Docker allocation, process limits, and whether executions were serial, interleaved, or concurrent not recorded} & \hl{Same workstation and deployment layer; host and container limits and execution scheduling not recorded} \\
\hl{Attempt and termination policy} & \hl{One scored attempt per prompt in the 100-prompt benchmark; crashes, hangs, failures, and timeouts retained as outcomes; timeout duration, retry policy, reset procedure, and detailed-layer attempt policy not recorded} & \hl{Same recorded primary-layer policy; timeout duration, retry policy, reset procedure, and detailed-layer attempt policy not recorded} \\
\hl{Wall-time measurement} & \hl{Benchmark-harness wall-clock timing from immediately before task execution until completion, failure, or termination; startup/readiness latency recorded separately; collector implementation not recorded} & \hl{Same timing boundary and reporting procedure} \\
\hl{Peak-memory measurement} & \hl{Container/runtime-level observed peak memory; collector, unit convention, and sampling method not recorded} & \hl{Same stated scope; collector, unit convention, and sampling method not recorded} \\
\hl{Outcome assessment} & \hl{Manual score: pass = 1, partial pass = 0.5, fail = 0. Pass required a correct and complete result; partial pass required a meaningful, verifiable task component or intermediate state. One non-blinded scorer; no independent second rating} & \hl{Same rubric and assessment procedure} \\
\hl{Execution period and lineage} & \hl{April--June 2026; timestamped run identifiers and CSV records; mapping between the 100-prompt and 23-prompt execution attempts not recorded} & \hl{Same period and record structure; mapping between evidence layers not recorded} \\
\bottomrule
\end{tabularx}
\end{table*}

\subsection{Subset composition and simultaneous-zero audit}\label{app:subset-audit}

\hl{The 23 prompt identifiers have a horizon composition close to the remaining 77 prompts, but their category coverage differs. The subset contains 7 short, 8 medium, and 8 long prompts; the remainder contains 23, 27, and 27. The subset omits Accuracy, Continuity, and Startup, includes the only Traceability prompt, and represents Auditability, Reliability, and Tool Use at higher shares. The supplementary table subset-representativeness.csv reports all category counts and proportions.}

\hl{Using primary-layer outcomes for the same identifiers, the OpenClaw subset contains 10 failures, 5 partial completions, and 8 passes, compared with 38, 16, and 23 in the remainder. The NanoBot subset contains 9 failures, 7 partial completions, and 7 passes, compared with 44, 15, and 18 in the remainder. These differences indicate that the detailed subset has a distinct outcome composition, particularly for NanoBot, and define the scope of its resource results.}

\hl{The ten simultaneous-zero operational rows comprise nine OpenClaw records and one NanoBot record, all associated with failure outcomes. Excluding the union of the affected prompt identifiers from both systems leaves 90 paired prompts and produces 30 versus 24 full completions, a risk difference of 0.067, and exact McNemar} $p=0.263$\hl{. Supplementary tables list the affected records and reproduce this sensitivity analysis.}

\subsection{Exploratory category-level effects}

\begin{figure*}[htbp]
\centering
\includegraphics[width=0.92\textwidth]{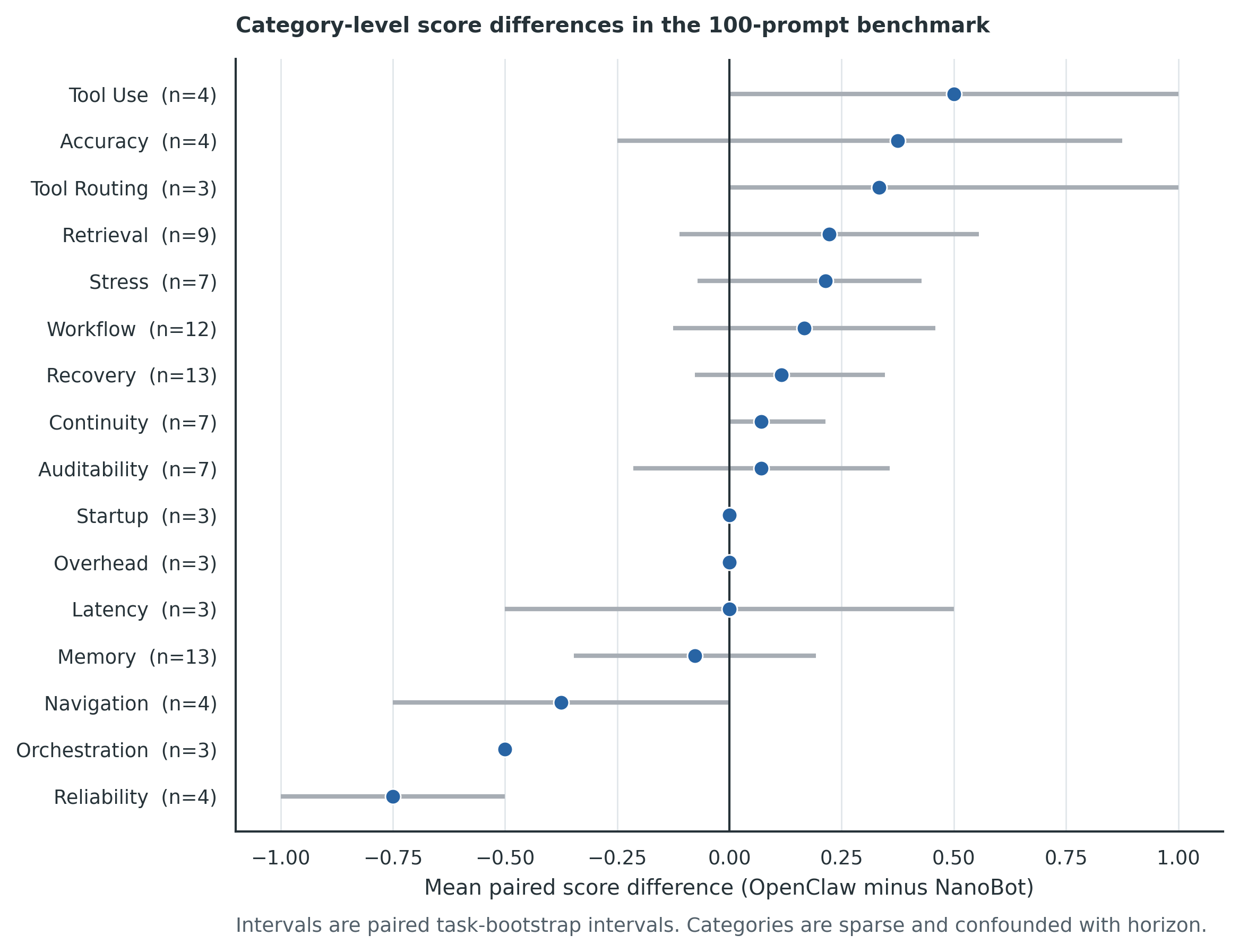}
\caption{Exploratory category-level paired score differences for the 16 categories containing at least three prompts. Traceability is omitted because it contains one prompt. Positive values favour OpenClaw and negative values favour NanoBot. Intervals use paired task-bootstrap resampling. No multiplicity adjustment is applied, and category membership overlaps with horizon and environment demand.}\label{fig:category-effects}
\end{figure*}

Figure~\ref{fig:category-effects} disaggregates the secondary ordinal score difference across the 16 categories containing at least three prompts. The Traceability category is omitted because it contains one prompt. Positive estimates favour OpenClaw and negative estimates favour NanoBot; prompt counts appear beside each category.

\hl{The estimates span from 0.50 for Tool Use to} $-0.75$ \hl{for Reliability. These extremes occur in categories with four prompts. Orchestration also favours NanoBot by} $-0.50$ \hl{across three prompts, whereas the two largest categories show smaller differences: 0.115 for Recovery and} $-0.077$ \hl{for Memory, each with 13 prompts. Sparse cells, wide intervals, and overlap between category and horizon frame these results as task-dependent patterns. The estimates identify task families for a balanced benchmark designed to test category-specific system strengths.}

\subsection{Capability-cost position}

\begin{figure*}[htbp]
\centering
\includegraphics[width=0.94\textwidth]{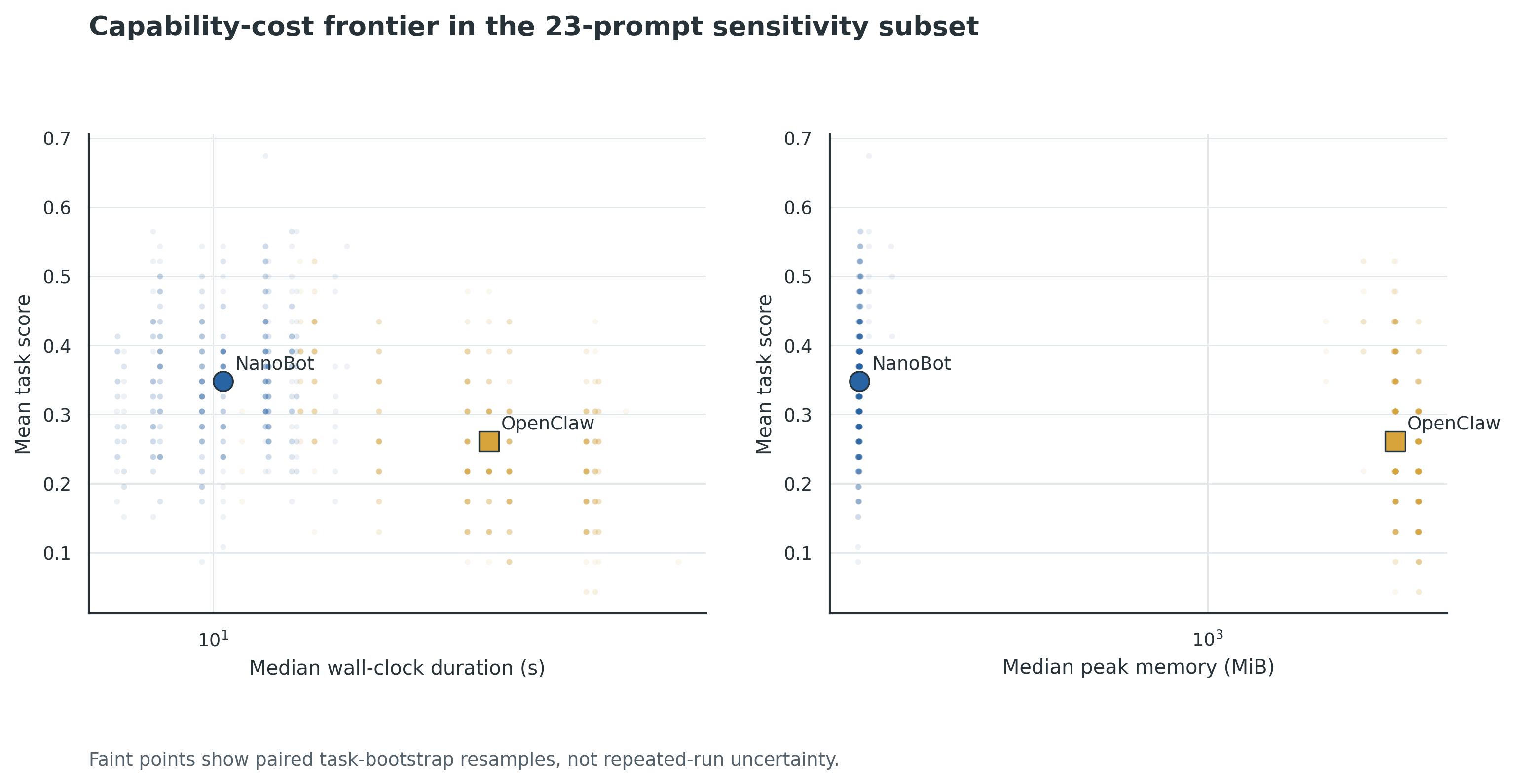}
\caption{Capability-cost position in the detailed execution subset using median wall-clock duration and peak memory. \protect\hl{Faint points show paired task-bootstrap resamples over prompts.}}\label{fig:capability-cost-frontier}
\end{figure*}

Figure~\ref{fig:capability-cost-frontier} places the mean ordinal outcome beside the median wall time and median peak memory in the detailed execution subset. It complements the prompt-level dominance analysis by showing the aggregate location of \hl{each system} on both resource axes.

NanoBot occupies the lower-cost, higher-score position in both panels: its mean score is 0.348 rather than 0.261, its median wall time is 10.446 rather than 34.063 seconds, and its median peak memory is 136.1 rather than 2926.6 MiB under the stated unit assumption. \hl{The bootstrap clouds quantify prompt-resampling variation across the 23 observed prompt pairs and characterise the capability-cost positions of the two systems within the detailed prompt set.}

\subsection{Cross-layer outcome concordance}

\begin{figure*}[htbp]
\centering
\includegraphics[width=0.90\textwidth]{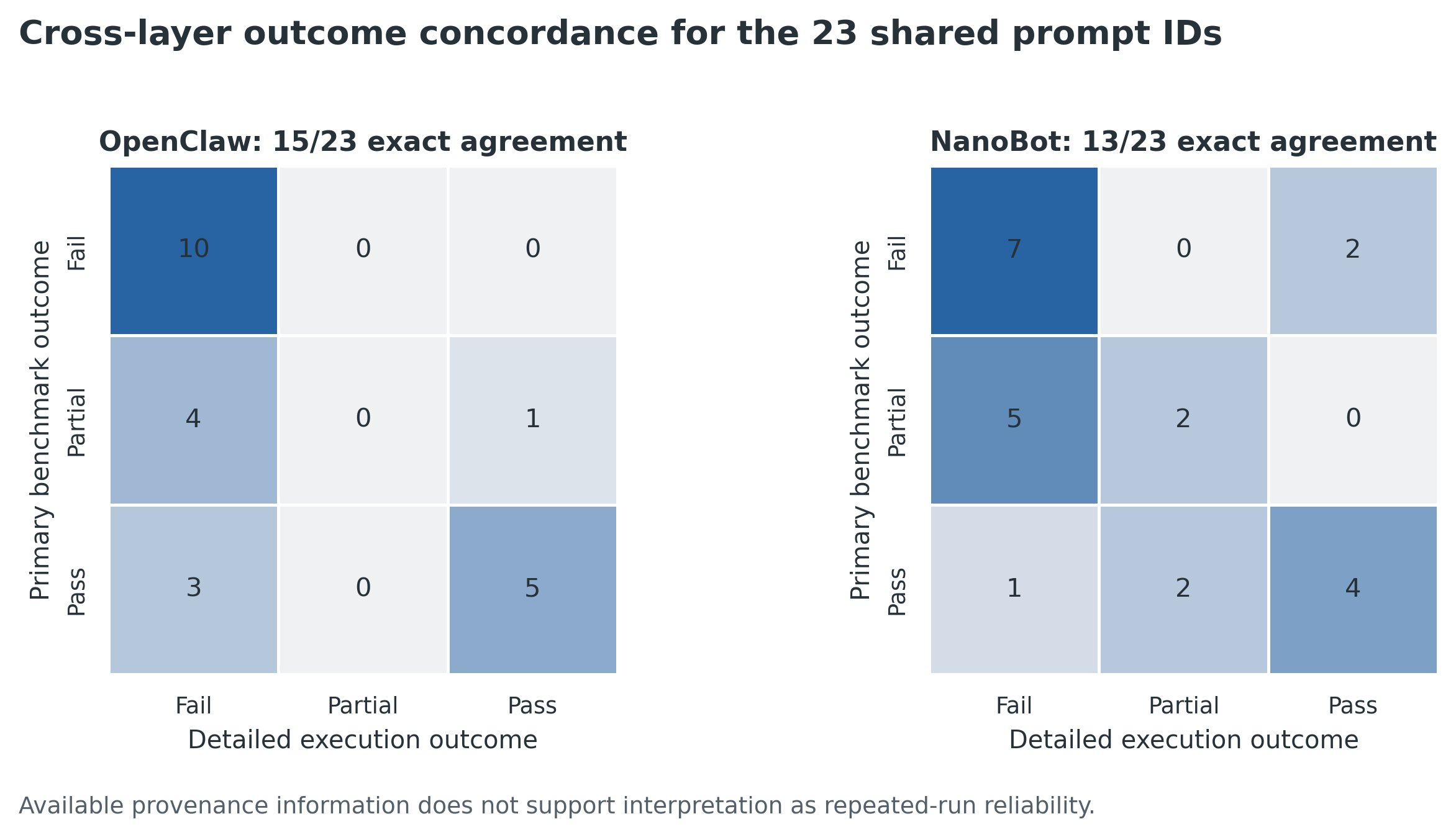}
\caption{Outcome correspondence between the primary benchmark and detailed execution records for 23 shared prompt identifiers. \protect\hl{The figure provides a cross-layer concordance audit; repeated-run analysis additionally requires shared attempt-level lineage.}}\label{fig:cross-layer}
\end{figure*}

Figure~\ref{fig:cross-layer} cross-tabulates the ordinal outcome assigned to each shared prompt in the primary benchmark and detailed execution layer. Diagonal cells denote exact agreement; off-diagonal cells identify changes in outcome category.

\hl{OpenClaw agrees across layers on 15 of 23 prompts. Its eight disagreements comprise four primary partial outcomes classified as failures in the detailed layer, one primary partial classified as a pass, and three primary passes classified as failures. NanoBot agrees on 13 prompts. Its ten disagreements include two primary failures classified as detailed-layer passes, five primary partial outcomes classified as failures, one primary pass classified as a failure, and two primary passes classified as partial outcomes.} The changes therefore do not form a uniform upward or downward shift. \hl{Rerunning, rescoring, environment change, or a combination could produce this pattern. The matrices demonstrate why attempt-level lineage is required before cross-layer differences can support reliability conclusions.}

\end{document}